\documentclass[air,twoside,11pt,theapa]{article}

\makeatletter
 \def\@textbottom{\vskip \z@ \@plus 1pt}
 \let\@texttop\relax
\makeatother

\usepackage[utf8]{inputenc}
\usepackage{graphicx}
\usepackage[usenames,dvipsnames]{xcolor}
\usepackage{lipsum}  
\usepackage{lineno}
\usepackage{booktabs}       % professional-quality tables
\usepackage{amsfonts}       % blackboard math symbols
\usepackage{nicefrac}       % compact symbols for 1/2, etc.
\usepackage{microtype}      % microtypography
\usepackage{siunitx}        % SI units

\usepackage{amssymb}
\usepackage{amsmath}
\DeclareMathOperator*{\argmax}{arg\,max}

\usepackage[font={small}]{subcaption}       % Error: Undefined package

\usepackage{wrapfig}
\usepackage{xspace}
\usepackage{pifont}
\usepackage{placeins}
\usepackage{paralist}
\usepackage[nolist,smaller]{acronym}
\usepackage{multirow}
\usepackage{tabularx}
\usepackage{ragged2e}
\usepackage{booktabs}

\usepackage{graphicx}
\usepackage{bm}
\usepackage{blindtext}
\usepackage{url}

\usepackage[capitalise,noabbrev]{cleveref}
\usepackage{tcolorbox}
\tcbuselibrary{most}
\usepackage{ltablex}
\usepackage{booktabs}
\usepackage{jair}
\usepackage{rawfonts}
\usepackage{theapa}

\jairheading{79}{2024}{1167-1236}{11/2023}{04/2024}
\ShortHeadings{Structure in Deep Reinforcement Learning: A Survey and Open Problems}
{Mohan, Zhang, \& Lindauer}
\firstpageno{1167}

\begin{document}
% !TEX root = main.tex
\newcommand{\RL}{Reinforcement Learning\xspace}
\newcommand{\rl}{reinforcement learning\xspace}
\newcommand{\mdp}{Markov Decision Process\xspace}
\newcommand{\pomdp}{Partially Observable MDP\xspace}

\newcommand{\fmdp}{Factored MDP\xspace}
\newcommand{\smdp}{Semi-MDP\xspace}
\newcommand{\bmdp}{Block MDP\xspace}
\newcommand{\rmdp}{Relatonal MDP\xspace}
\newcommand{\oomdp}{Object-Oriented MDP\xspace}

\definecolor{my_magenta}{rgb}{0.796875,0.46875,0.734375}
\definecolor{my_green}{rgb}{0.0078125,0.6171875,0.44921875}
\definecolor{my_gold}{rgb}{0.8671875,0.55859375,0.01953125}
\definecolor{my_gray}{rgb}{0.5,0.5,0.5}
\definecolor{my_blue}{rgb}{0.4,0.4,1.}
\definecolor{my_green2}{HTML}{63a76b}
\definecolor{blue}{HTML}{526fae}

% Colors for the usage
\newcommand*{\sef}{\textcolor{Mahogany}}                 % Sample Efficiency
\newcommand*{\gen}{\textcolor{OliveGreen}}             % Generalization
\newcommand*{\safe}{\textcolor{blue}}               % Safety
\newcommand*{\inter}{\textcolor{Yellow}}           % Interpretability
\newcommand*{\explore}{\textcolor{RoyalPurple}}               % Exploration

\newcolumntype{Y}{>{\centering\arraybackslash}X}
\setlength\doublerulesep{0.4pt}

\newcommand\todo[1]{{\color{red}[ToDo: #1]}}

\newtcolorbox{mybox}{enhanced,colback=blue!5!white,colframe=blue!75!black,boxrule=0.6mm,fontupper=\footnotesize}

\title{Structure in Deep Reinforcement Learning:\\ A Survey and Open Problems}

\author{\name Aditya Mohan \email a.mohan@ai.uni-hannover.de \\
         \addr Institute of Artificial Intelligence\\
        Leibniz University Hannover\\ \\
        \name Amy Zhang \email amy.zhang@austin.utexas.edu \\ 
        \addr University of Texas at Austin,\\ Meta AI\\ \\
        \name Marius Lindauer \email m.lindauer@ai.uni-hannover.de \\
        \addr Institute of Artificial Intelligence,\\ L3S Research Center\\
        Leibniz University Hannover\\
}

\maketitle

\begin{abstract}
Reinforcement Learning (RL), bolstered by the expressive capabilities of Deep Neural Networks (DNNs) for function approximation, has demonstrated considerable success in numerous applications. 
However, its practicality in addressing various real-world scenarios, characterized by diverse and unpredictable dynamics, noisy signals, and large state and action spaces, remains limited. 
This limitation stems from poor data efficiency, limited generalization capabilities, a lack of safety guarantees, and the absence of interpretability, among other factors.
To overcome these challenges and improve performance across these crucial metrics, one promising avenue is to incorporate additional structural information about the problem into the RL learning process.
Various sub-fields of RL have proposed methods for incorporating such inductive biases.
We amalgamate these diverse methodologies under a unified framework, shedding light on the role of structure in the learning problem, and classify these methods into distinct patterns of incorporating structure.
By leveraging this comprehensive framework, we provide valuable insights into the challenges of structured RL and lay the groundwork for a design pattern perspective on RL research. 
This novel perspective paves the way for future advancements and aids in developing more effective and efficient RL algorithms that can potentially handle real-world scenarios better.
\end{abstract}
\section{Introduction}

%While RL has been successful, it is not deployable 
\RL (RL) has contributed to a range of sequential decision-making and control problems like games~\shortcite{silver-nature16a}, robotic manipulation~\shortcite{lee-sciro20}, and optimizing chemical reactions~\shortcite{zhou-acs17a}.
%To tackle deployability, we need to talk about the generalization problem 
Most of the traditional research in RL focuses on designing agents that learn to solve a sequential decision problem induced by the inherent dynamics of a task, e.g., the differential equations governing the cart pole task~\shortcite{sutton-book18b} in the classic control suite of OpenAI Gym~\shortcite{brockman-arxiv16}. 
However, their performance significantly degrades when even minor aspects of the environment change~\shortcite{meng-data19,lu-amia20}. 
Moreover, deploying RL agents for real-world learning-based optimization has additional challenges, such as complicated dynamics, intractable and computationally expensive state and action spaces, and noisy reward signals.

Thus, research in RL has started to address these issues through methods that can generally be categorized into two dogmas~\shortcite{mannor-arxiv23a}:
\begin{inparaenum}[(i)]
    \item \textbf{Generalization:} Methods developed to solve a broader class of problems where the agent is trained on various tasks and environments~\shortcite{kirk-jair23b,benjamins-tmlr23a}. 
    \item \textbf{Deployability:} Methods specifically engineered towards concrete real-world problems by incorporating additional aspects such as feature engineering, computational budget optimization, and safety. 
\end{inparaenum}
% How does structure come into this picture
The intersection of generalization and deployability covers problems requiring methods to handle sufficient diversity in the task while being deployable for specific applications. 
To foster research in this area, \shortciteA{mannor-arxiv23a} argue for a design-pattern oriented approach, where methods can be abstracted into patterns that are specialized to specific problems. 

\begin{figure}[ht]
    \centering
    \includegraphics[width=\textwidth]{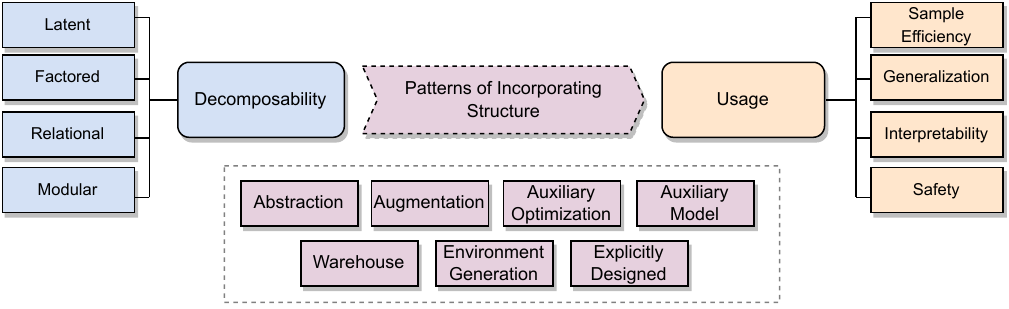}
    \caption{\textbf{Overview of our framework.}
    % We discuss three axes of Structure in Reinforcement Learning: Usage, Decomposition, and Patterns. 
    Side information can be used to achieve improved performance across metrics such as \emph{Sample Efficiency}, \emph{Generalization}, \emph{Interpretability}, and \emph{Safety}. We discuss this process in \cref{sec:usage}.
    A particular source of side information is decomposability in a learning problem, which can be categorized into four archetypes along a spectrum - \emph{Latent}, \emph{Factored}, \emph{Relational}, and \emph{Modular} - explained further in \cref{sec:Structure:types}.
    Incorporating side information about decomposability amounts to adding structure to a learning pipeline, and this process can be categorized into seven different patterns - \emph{Abstraction}, \emph{Augmentation}, \emph{Auxiliary Optimization}, \emph{Auxiliary Model}, \emph{Warehouse}, \emph{Environment Generation}, and \emph{Explicitly Designed} - discussed further in \cref{sec:patterns}.}
    \label{fig:structure-overview}
\end{figure}

%Pitch
However, the path to RL design patterns is hindered by gaps in our understanding of the relationship between the design decisions for RL methods and the properties of the problems they might be suited for. 
While decisions like using state abstractions for high-dimensional spaces seem obvious, decisions like using relational neural architectures for problems are not so apparent to a designer. 
One way to add principle to this process is to understand how to incorporate additional domain knowledge into the learning pipeline. 
The structure of the learning problem itself, including priors about the state space, the action space, the reward function, or the dynamics of the environment, is a vital source of such domain knowledge.
While such methods have been research subjects throughout the history of RL~\shortcite{parr-neurips97}, approaches that try to achieve this in Deep RL are scattered across the various sub-fields in the vast and disparate landscape of modern RL research.
In this work, we take the first steps to amalgamate these approaches under our pattern-centric framework for incorporating structure in RL.
\cref{fig:structure-overview} shows a general overview of three elements of understanding the role of incorporating structure into a learning problem that we cover in this work. 

We refer to side information as additional knowledge not required to formulate the vanilla MDP.
Incorporating structure entails utilizing side information about decomposability to improve sample efficiency, generalization, interpretability, and/or safety. 
To build an intuition about what we mean by this, consider the task of tailoring educational content to individual learners based on their preferences, learning pace, and mastery levels.
An RL agent can select appropriate learning materials, activities, and assessments that best fit a learner's current state and learning goals. 
Such a scenario is rich with structural properties and decompositions, such as learning styles or hidden skill proficiency of learners, distinct areas of knowledge within a learning program, relationships between knowledge areas and skills of learners, and modular content delivery mechanisms.
While an MDP can potentially be formulated by treating the problem as a monolith, it does not need to be the most efficient solution. 
Instead, the problem can be formulated in various ways where prior knowledge about such decomposability can encode inductive biases into the RL agent. 
Prior knowledge about decompositions can additionally be discovered through auxiliary methods, such as Large Language Models (LLMs), that can analyze vast amounts of educational content, extracting key concepts, learning objectives, and difficulty levels.
Incorporating side information into the learning pipeline, such as using the LLM to generate an intrinsic reward~\shortcite{klissarov-iclr24a}, can improve the speed of convergence of the RL agent, make it robust to variations in the problem and potentially help with making it safer and more interpretable.

% To build an intuition about what we mean by this, consider the task of managing a large factory with many production cells~\footnote{example taken from \shortciteA{guestrin-jair03a}}. 
% If a cell positioned early in the production line generates faulty parts, the whole factory may be affected. 
% However, the quality of the parts a cell generates depends directly only on the state of this cell and the quality of the parts it receives from neighboring cells. 
% Additionally, the cost of running the factory depends, among other things, on the sum of the costs of maintaining each local cell. 
% Finally, while a cell responsible for anodization may receive parts directly from any other cell in the factory, a work order for a cylindrical part may restrict this dependency to cells with a lathe. 
% In the context of producing cylindrical parts, the quality of the anodized parts depends only on the state of cells with a lathe.  
% Thus, by incorporating information about the additive nature of production, costs, and the context of the part that needs to be produced, the learning pipeline can be imbued with better objectives such as improved sample efficiency or robustness of a learned policy to changing factory conditions.
% \end{rev-C}
% The vanilla MDP framework does not require incorporating such additive information into the learning pipeline. 
% Thus, biasing the pipeline with this amounts to incorporating structure.

\paragraph{Structure of the Paper.}
To better guide the reader, the paper is structured as follows: \begin{inparaenum}[(i)]
    \item In \cref{sec:related-work} we discuss the related works. We cover previous surveys on different areas in RL and previous works aimed at incorporating domain knowledge into RL methods.
    \item In \cref{sec:preliminaries}, we describe the background and notation needed to formalize the relevant aspects of the RL problems. We additionally define the RL pipeline that we use in the later sections.
    \item In \cref{sec:usage}, we introduce side information and define the additional metrics that can be addressed by incorporating side information into an RL pipeline. 
    \item In \cref{sec:Structure}, we formulate structure as side information about decomposability and categorize decompositions in the literature into four archetypes on the spectrum of decomposability~\shortcite{hofer-book17a}. Using these archetypes, we demonstrate how various problem formulations in RL fall into the proposed framework. 
    \item In \cref{sec:patterns}, we formulate seven patterns of incorporating structure into the RL learning process and provide an overview of each pattern by connecting it to the relevant surveyed literature. We represent each pattern graphically as a plug-and-play modification to the RL pipeline introduced in \cref{sec:preliminaries}. We additionally provide a literature survey for each pattern as a table and show possible research areas as empty spaces.
    \item In \cref{sec:connection}, we discuss how our framework opens new avenues for research while providing a common reference point for understanding what kind of design decisions work under which situations. We additionally summarize concrete takeaways for researchers and practitioners in various research areas in RL.
\end{inparaenum}

\paragraph{Scope of the Work.}
Using structure has had a long history in RL, with early ideas already surfacing in the classical RL literature~\shortcite{boutilier-ijcai95,fitch-book05a,sutton-book18b}. 
Given the vast nature of this class of methods, we limit our focus in three ways:
\begin{inparaenum}[(i)]
    \item we primarily cover Deep RL methods developed in the last ten years. While we discuss earlier works to establish the traditional underpinnings of our conceptual framework in \cref{sec:Structure}, our survey in \cref{sec:patterns} only covers the later Deep RL literature;
    \item we only cover single-agent RL in this work. Multi-agent RL (MARL)~\shortcite{gronauer-air22a} provides an additional dimension of decomposability, enabling additional patterns. While mathematically, certain aspects of such settings can be modeled equivalently by subsuming certain notions into the single-agent RL framework, we consider this field to have sufficient nuance and complexity to deserve a separate in-depth analysis and
    \item we do not cover theoretical work related to structure in RL. Such methods study how incorporating additional information and decompositions affect metrics such as learning complexity~\shortcite{sun-colt19,agarwal-neurips20}, and could be potentially categorized into our suggested framework. However, we focus on empirical research because it is more widely studied. 
\end{inparaenum}

\section{Related Work}
\label{sec:related-work}

Multiple surveys have previously covered different areas in RL.
However, none have covered the methods of explicitly and holistically incorporating structure in RL.
In the following sections, we divide our literature research into surveys that tackle different problem settings, additional objectives, individual decompositions, and previous works incorporating domain knowledge into RL pipelines.

\paragraph{Different RL settings.}
\shortciteA{kirk-jair23b} survey the field of Zero-Shot generalization and briefly discuss the need for more restrictive structured assumptions for their setting.
While their survey argues for the requirement of similar assumptions, our work specifically lays out a framework that allows surveying approaches to utilize these assumptions. 
Additionally, our work is not limited to the setting of zero-shot generalization but covers additional areas of interpretability, safety, and sample efficiency in RL.
\shortciteA{beck-arxiv23a} cover the field of Meta-RL and discuss the role of structure in Meta-Exploration, Transfer, and the POMDP formulation of Meta-Learning.
However, their focus is on surveying the Meta-Learning setting and does not delve deeper into grounding what structure means, as is the case with our work.
This is also the case with the survey of exploration methods in RL~\shortcite{amin-corr21}, where they argue for the need to choose the policy space to reflect prior information about the structure of the solution to ensure that the exploration behavior follows the same structure. 
Our framework grounds this idea in decomposition and argues for incorporating this information using one of many patterns.

\paragraph{Additional objectives.}
Individual surveys have additionally covered the multiple objectives defined in \cref{sec:usage}.
\shortciteA{garcia-jmlr15} provides a comprehensive review of the literature on safety in RL and divides the methods based on whether they modify the optimization criterion or the exploration process. 
We use their categorization to examine the correlation of patterns that use structural information for safety but cover additional objectives beyond safety. 
In a similar vein, \shortciteA{glanois-arxiv21a} cover methods that add interpretability to the RL pipeline, which judges interpretability along the same axis as our work, namely, the definitions proposed by \shortciteA{lipton-acm18}.

\paragraph{Grounding decompositions.}
The assumptions of decomposability proposed in \cref{sec:Structure:types} utilize the spectrum of decomposability previously proposed by \shortciteA{hofer-book17a}.
We add to this framework by pinpointing four major decomposability archetypes on this spectrum, allowing us to build our categorization framework.
Previous surveys have covered ideas related to these archetypes individually.
\shortciteA{pateria-acmcs22a} survey hierarchical methods that come under modular decompositions in our framework.
\shortciteA{zhang-aimag22a} survey methods that leverage reasoning and declarative knowledge for sequential decision-making, including RL. 
A class of such methods falls under the relational decompositions in our framework.

\paragraph{Incorporating domain knowledge into RL.}
Certain surveys have also been conducted on methods incorporating domain knowledge into RL. 
\shortciteA{eser-ras22a} survey methods that incorporate additional knowledge to tackle real-world deployment in RL. 
To this end, they categorize sources of knowledge into three types: 
\begin{inparaenum}[(i)]
    \item Scientific Knowledge, that covers empirical knowledge about the problem;
    \item World Knowledge, that covers an intuitive understanding of the problem that can be incorporated into the pipeline; and,
    \item Expert knowledge, available to experienced professionals in the form of experience.
\end{inparaenum}
They formalize an RL pipeline and then look at methods incorporating this knowledge into different parts of the pipeline, such as problem representation, learning strategy, task structuring, and Sim2Real transfer. 
In addition to our scope focusing on domain knowledge about decomposability, our approach is source-agnostic.
We focus on the specific part of the MDP on which structural assumptions are imposed and the nature of such assumptions. 
We categorize methods into patterns of incorporating these problems' assumptions. 
Additionally, our patterns framework covers a broader range of methods that apply to more settings than Sim2Real.

The intersection of side information and patterns has previously been discussed by \shortciteA{jonschkowski-arxiv15} and inspires our categorization as well.
However, they predominantly discuss patterns for supervised and semi-supervised settings and mention trivial extensions to state representations in RL. 
Our formulation of patterns covers the RL pipeline more holistically by additionally looking at assumptions on components such as actions, transition dynamics, learned models, and previously learned skills. 
Moreover, our formulation solely focuses on different ways of biasing RL pipelines, holding little relevance for supervised and semi-supervised learning communities. 

\section{Preliminaries}
\label{sec:preliminaries}

The following sections summarize the main background necessary for our approach to studying structural decompositions and related patterns. 
In \cref{sec:preliminaries:MDP}, we formalize the sequential decision-making problem as an MDP.
\cref{sec:preliminaries:RL} then presents the RL framework for solving MDPs and introduces the RL pipeline.

\subsection{Markov Decision Processes}
\label{sec:preliminaries:MDP}

Sequential decision-making problems are usually formalized using the notion of a \mdp (MDP) \shortcite{bellman-1954,puterman-2014}, which can be written down as a 5-tuple $\mathcal{M} = \langle \mathcal{S}, \mathcal{A}, R, P, \rho \rangle$. 
At any timestep, the environment exists in a state $s  \in \mathcal{S}$, with $\rho$ being the initial state distribution.
The agent takes an action $a \in \mathcal{A}$ which \emph{transitions} the environment to a new state $s'  \in \mathcal{S}$. 
The stochastic transition function governs the dynamics of such transitions $P: \mathcal{S} \times  \mathcal{A} \to \Delta(\mathcal{S})$, which takes the state $s$ and action $a$ as input and outputs a probability distribution over the following states $\Delta(.)$ from which the subsequent state $s'$ can be sampled.
For each transition, the agent receives a reward $R: \mathcal{S} \times \mathcal{A} \to \mathbb{R}$, with $R \in \mathcal{R}$. 
The sequence $(s, a, r, s')$ is an experience.

The agent acts according to a policy  $\pi: \mathcal{S} \to  \Delta(\mathcal{A})$, in a space of policies $\Pi$, that produces a probability distribution over actions given a state.
This is a delta distribution for deterministic policies, which leads to the policy outputting a single action.  
Using the current policy, an agent can repeatedly generate experiences, and a sequence of such experiences is also called a \emph{trajectory} ($\tau$):

\[
    \tau  = \{(s_t, a_t, r_t, s_{t+1} )\}_{t\in [t_0, t_{T-1}]}  \,\,\,\, \forall (s,a,r, s) \in \mathcal{S} \times \mathcal{A} \times R \times \mathcal{S}.
\]

In episodic RL, the trajectory consists of experiences collected over multiple episodes with environment resets. 
In contrast, in continual settings, the trajectory encompasses experiences collected over some horizon in a single episode.
The rewards in $\tau$ can be accumulated into an expected sum called the return $G$, which can be calculated for any starting state $s$ as

\begin{equation}\label{eq:state-return}
      G(\pi, s) = \mathbb{E}_{(s_0 = s, a_1, r_1, \dots) \sim \pi} \bigg [ \sum_{t=0}^{\infty}  r_t \bigg ].
\end{equation}

\noindent
For the sum in \cref{eq:state-return} to be tractable, we either assume the horizon of the problem to be of a fixed length $T$ (finite-horizon return), i.e., the trajectory to terminate after $T$-steps, or we discount the future rewards by a discount factor $\gamma$ (infinite horizon return). 
Discounting, however, can also be applied to finite horizons. 
Solving an MDP amounts to determining the policy $\pi^* \in \Pi$ that maximizes the expectation over the returns of its trajectory. 
This expectation can be captured by the (state-action) value function $Q \in \mathcal{Q}$. 
Given a policy $\pi$, the expectation can be written recursively:

% \begin{equation}
%     Q^{\pi} (s,a) = \mathbb{E}_{s \sim \rho } \big[ G_t \mid s, a \big] =  \mathbb{E}_{s' \sim \tau} \big [ R(s,a)  +  \gamma \mathbb{E}_{a' \sim \pi(\cdot \mid s')} [Q^{\pi}(s', a') ] \big ].
% \end{equation}

\begin{equation}
    Q^{\pi} (s,a) = \mathbb{E}_{\pi} \big[ \sum_{t=0}^T r_t \mid s_0=s, a_0=a \big] =  \mathbb{E}_{\pi} \big [ R(s,a)  +  \gamma \mathbb{E}_{a' \sim \pi(\cdot \mid s')} [Q^{\pi}(s', a') ] \big ].
\end{equation}

\noindent
Thus, the goal can now be formulated as the task of finding an optimal policy that can maximize the $Q^{\pi} (s,a)$:

\begin{equation}\label{eq:optimal-policy}
    \pi^* \in \argmax_{\pi \in \Pi} Q^{\pi}(s,a) \,\,\,\, \forall (s,a) \in \mathcal{S} \times \mathcal{A}.
\end{equation}

\noindent
We also consider Partially Observable MDPs (POMDPs), which model situations where the state is not fully observable. 
A POMDP is defined as a 7-tuple $\mathcal{M} = \langle \mathcal{S}, \mathcal{A}, \mathcal{O}, R, P, \xi, \rho \rangle$, where $\mathcal{S}, \mathcal{A}, R, P, \rho$ remain the same as defined above.
Instead of observing the state $s \in \mathcal{S}$, the agent now has access to observation $o \in \mathcal{O}$ that is generated from the actual state through an emission function $\xi: \mathcal{S} \times \mathcal{A} \to \Delta(\mathcal{O})$. 
Thus, the observation takes the state's role in the experience generation process.
However, solving POMDPs requires maintaining an additional belief since multiple $(s,a)$ can lead to the same $o$.

\subsection{Reinforcement Learning}
\label{sec:preliminaries:RL}
 
The task of an RL algorithm is to interact with the MDP by simulating its transition dynamics $P(s'\mid s,a)$ and reward function $R(s, a)$ and learn the optimal policy mentioned in \cref{eq:optimal-policy}.
In Deep RL, the policy is a Deep Neural Network~\shortcite{goodfellow-16a} that is used to generate $\tau$. 
We can optimize such a policy by minimizing an appropriate objective $J$.

\begin{figure}[htb]
    \centering
    \includegraphics[width=\textwidth]{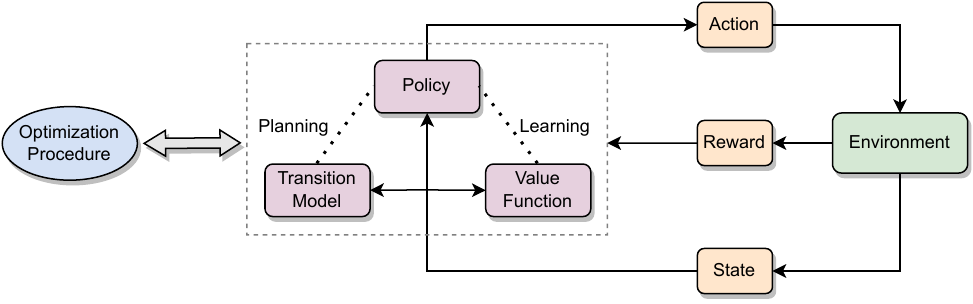}
    \caption{The anatomy of an RL pipeline.}
    \label{fig:rl-pipeline}
\end{figure}

A model of an MDP $\hat{\mathcal{M}}$ allows an agent to \emph{plan} a trajectory by simulating it to generate experiences.
RL methods that use such models are categorized into \emph{Model-Based RL} \shortcite{moerland-ftml23a}. 
On the other hand, not having such a model necessitates learning the policy directly from experiences, and such methods fall into the category of \emph{Model-free RL}.

RL methods can additionally be categorized based on the type of objective $J$. 
Methods that use a value function, and correspondingly either Monte-Carlo estimates or Temporal Difference (TD) error~\shortcite{sutton-ml88}, to learn a policy fall into the category of \emph{Value-based RL}. 
A key idea in TD methods is \emph{bootstrapping}, where they use a learned value estimate to improve the estimate of a state that precedes it.
\emph{On-policy} methods directly update the policy that generated the experiences, while \emph{Off-policy} methods use a separate policy to generate experiences. 
\emph{Policy-Based} Methods parameterize the policy directly and use the policy gradient theorem~\shortcite{williams-ML92,sutton-neurips99} to create $J$. 

A central research theme in practical RL methods focuses on approximating a global solution by iteratively learning one or more of the aforementioned quantities using supervised learning and function approximations. 
% Deep RL~\shortcite{franccois-ftml18} focuses explicitly on using Deep Neural Networks \shortcite{goodfellow-16a} as function approximators and has shown state-of-the-art (SOTA) performance on various tasks.
We use the notion of a pipeline to talk about different RL methods.
\cref{fig:rl-pipeline} shows the anatomy of an RL pipeline.
The pipeline can be defined as a mathematical tuple $\Omega = \langle \mathcal{S}, \mathcal{A}, R, P, Q, \pi, \hat{\mathcal{M}}, J, \mathcal{E} \rangle$, where all definitions remain the same as before.  
To solve an MDP, the pipeline operates on given an environment $\mathcal{E}$ by taking the state $s \in \mathcal{S} $ as input and producing an action $ a \in \mathcal{A}$ as an output. The environment operates with the dynamics $P$ and a reward function $R$. 
The pipeline might generate experiences by directly interacting with $\mathcal{E}$, i.e., \emph{learning} from experiences or by simulating a learned model $\hat{\mathcal{M}}$ of the environment.
The optimization procedure encompasses the interplay between the current policy $\pi$, its value function $Q$, the reward $R$, and the learning objective $J$. 
% With a slight abuse of notation, we refer to any of the components of a pipeline as $X$ and assume the space in which it exists as $\mathcal{X}$.

\section{Side Information and its Usage}
\label{sec:usage}

In addition to the characterization of the problem by an MDP, there can still be additional information that could potentially improve performance on additional metrics such as \emph{Sample Efficiency}, \emph{Generalization}, \emph{Interpretability}, and \emph{Safety}.
We call this \emph{Side Information} (also called privileged information). 
For the (semi-) supervised and unsupervised settings, side information is any additional information that, while neither part of the input nor the output space, can potentially contribute to the learning process \shortcite{jonschkowski-arxiv15}. 

Translated to the RL setting, this can be understood as additional information not provided in the original MDP definition $\mathcal{M}$.
Side information can be incorporated into the RL pipeline by biasing one or more components shown in \cref{fig:rl-pipeline}.
Mathematically, we can express this using some function $\beta$ that conditions the pipeline on side information by augmenting it with a function $Z$.

\[
    \beta: \Omega \to \Omega \times \mathcal{Z}
\]

This implies that we now augment our tuple $\Omega$ with an additional function $Z$ that operates on other tuple elements $\mathcal{X} \in \Omega$. 
For example, incorporating side information could be used to learn state abstractions by adding an encoder $Z$ to map the state space $\mathcal{S}$ to a latent representation $\kappa$ that can be used for control.
We discuss the general templates for $Z$ in \cref{sec:Structure} and classify different methods of biasing $\Omega$ with $Z$ into patterns in  \cref{sec:patterns}.
The natural follow-up question, then, becomes the impact of incorporating side information into the learning pipeline.
In this work, we focus on four ways side information can be used and formally define them in the following sections.

\subsection{Sample Efficiency} 
Sample Efficiency is intimately tied to the Sample Complexity of RL. 
Intuitively, if a pipeline demonstrates a higher reward than a baseline for the same number of timesteps, we consider it more sample-efficient.
To formally define it, we use the notion of the \emph{Sample Complexity of Exploration} \shortcite{kakade-misc03a,strehl-jmlr09a}: 
Given some $\epsilon > 0$, we can define the sample complexity as the number of timesteps  $t$ after which the policy produces a value $V^\pi < V^\star - \epsilon$.
This definition by \shortciteA{strehl-jmlr09a} directly measures the number of times the agent acts poorly (quantified by $\epsilon$) and views “fast” learners as those that act poorly as few times as possible.
Incorporating side information leads to a reduction in sample complexity, thus improving the sample efficiency.
% Intuitively, if a pipeline demonstrates a higher reward than a baseline for the same number of timesteps, we consider it more sample-efficient.
% However, methods can make specific claims on $\zeta$ by utilizing certain assumptions about the problem itself~\shortcite{deramo-iclr20a,modi-atl18a,brunskill-uai13a}.  

\paragraph{Exploration.}
One specific way to improve the sample complexity of exploration is to directly impact the exploration mechanism using side information. 
\shortciteA{amin-corr21} categorize exploration methods based on the type of information that an agent uses to explore the world into the following categories: 
\begin{inparaenum}[(i)]
    \item \emph{Reward-Free Exploration} methods in which extrinsic rewards do not affect the choice of action. Instead, they rely on intrinsically motivated forms of exploration, such as diversity maximization~\shortcite{eysenbach-iclr19a}.
    \item \emph{Randomized Action Selection} methods use estimated value functions, policies, or rewards to induce exploratory behavior.
    \item \emph{Optimism/Bonus-Based Exploration} methods use the \emph{optimism in the face of uncertainty} paradigm to prefer actions with higher uncertain values.
    \item \emph{Deliberate Exploration} methods that either use posterior distributions over dynamics (Bayesian setup) or meta-learning techniques to optimally solve exploration and
    \item \emph{Probability Matching} methods that use heuristics to select the next action.
\end{inparaenum}
Incorporating side information into any of these methods generally improved the state-space coverage of the exploration mechanism.
We specifically cover methods that impact the exploration mechanism to improve sample efficiency and/or generalization.

\subsection{Transfer and Generalization}

Transfer and generalization encompass performance metrics that measure how an RL agent performs on a set of different MDPs:  
Transfer evaluates how well an agent, trained on some MDP $\mathcal{M}_i$, performs on another MDP $\mathcal{M}_j$. 
This can be either done in a zero-shot manner, where the agent is not fine-tuned on  $\mathcal{M}_j$, or in a few-shot manner, where the agent gets to make some policy updates on $\mathcal{M}_j$ to learn as fast as possible. Generally, the performance gap between the two MDPs determines the transfer performance.

\begin{equation}
    J_{\text{transfer}}(\pi) := \bm{G}(\pi, \mathcal{M}_{i} ) - \bm{G}(\pi, \mathcal{M}_{j} ).
\end{equation}

% Given a distribution over MDPs $p(\bm{\mathcal{M}})$, generalization extends this idea to training an agent on a subset $\bm{\mathcal{M}}_{train} \sim p(\mathcal{M}) $ and then testing its performance on a separate set of MDPs $\bm{\mathcal{M}}_{test} \sim p(\mathcal{M})$. 
% Consequently, the generalization objective~\shortcite{kirk-jair23b} that needs to be minimized becomes 

% \begin{equation}
%     \text{Gen}(\pi) := \bm{G}(\pi, \bm{\mathcal{M}}_{train} ) - \bm{G}(\pi, \bm{\mathcal{M}}_{test} ),
% \end{equation}
 
Generalization extends this idea to training an agent on a set of training MDPs $\bm{\mathcal{M}}_{train}$ and then evaluating its performance on a separate set of MDPs $\bm{\mathcal{M}}_{test}$. 
Consequently, the metric can measure generalization~\shortcite{kirk-jair23b}. 

\begin{equation}
    \text{Gen}(\pi) := \bm{G}(\pi, \bm{\mathcal{M}}_{train} ) - \bm{G}(\pi, \bm{\mathcal{M}}_{test} ).
\end{equation}

A more restrictive form of generalization can be evaluated when the training and testing MDPs are sampled from the same distribution, i.e., $\bm{\mathcal{M}}_{train}, \bm{\mathcal{M}}_{test} \sim p(\bm{\mathcal{M}})$.
Depending on how the transfer is done (zero-shot, few-shot, etc.), this notion covers any form of distribution of MDPs, including multi-task settings.
Incorporating side information into the learning can minimize $\text{Gen}(\pi)$. 
As argued by \shortciteA{kirk-jair23b}, we outline three manners for doing so for the zero-shot case:
\begin{inparaenum}[(i)]
    \item Increasing similarity between $ \bm{\mathcal{M}}_{train}$ and $\bm{\mathcal{M}}_{test}$ through techniques such as Data Augmentation, Domain Randomization, Environment Generation, or by implicitly or explicitly impacting optimization objectives;
    \item Handling differences between $ \bm{\mathcal{M}}_{train}$ and $\bm{\mathcal{M}}_{test}$ by encoding inductive biases, regularization, learning invariances, or online adaptation; and,
    \item Handling RL-specific issues such as exploration and non-stationary data-distributions. 
\end{inparaenum}

Our patterns framework in \cref{sec:patterns} proposes a new way to categorize these approaches into design patterns. 
Consequently, we cover all such forms of handling generalization.

% Generally, the literature uses structure in specific ways to improve generalization.
% For example, latent and modular decompositions have been used to learn subspaces of dynamics between morphologically different robotic hands. 
% Similarly, relational decomposition allows generalization to different entity and object relations. 

\subsection{Interpretability}
Interpretability refers to a mechanistic understanding of a system to make it more transparent. 
\shortciteA{lipton-acm18} enumerate three fundamental properties of model interpretability:
\begin{inparaenum}[(i)]
    \item \textbf{Simulatability} refers to the ability of a human to simulate the inner workings of a system,
    \item \textbf{Decomposability} refers to adding intuitive understanding to individual working parts of a system,
    \item \textbf{Transparency} refers to improving the understanding of a system's function (such as quantifying its convergence properties).
\end{inparaenum}

Given the coupled nature of individual parts of an RL pipeline, adding interpretability amounts to learning a policy for the MDP that adheres to at least one of multiple such properties.
Incorporating side information can help improve performance in all three aspects, depending on the nature of side information and what it encompasses.
However, we do not explicitly provide a formal metric for interpretability due to the potentially subjective nature of such metrics, especially in the case of RL, where performances on such metrics might differ depending on the environment.
We instead look for interpretability through the lens of decomposability, which we discuss further in \cref{sec:Structure}, by specifically checking for whether the decompositions leveraged by the methods are individually simulatable or add transparency to the action-selection mechanism.
% We consider claims on interpretability based on whether a given work additionally addresses at least two of these metrics.

\subsection{Safety}
Safety refers to learning policies that maximize the expectation of the return in problems in which it is important to ensure reasonable
system performance and/or respect safety-related constraints during the learning and/or deployment processes.
For example, Model-based RL methods usually learn a model of the environment and then use it to plan a sequence of actions.
However, such models are often learned from noisy data, and deploying them in the real world might lead an agent to catastrophic states.
Therefore, methods in the Safe-RL literature focus on incorporating safety-related constraints into the training process to mitigate such issues.

While Safety in RL is a vast field in and of itself~\shortcite{garcia-jmlr15}, we consider two specific categories in this work: \emph{Safe Learning with constraints} and \emph{Safe Exploration}.
The former subjects the learning process to one or more constraints $c_i \in C$~\shortcite{altman-book99a}.
Depending on the necessity of strictness, these can be incorporated in many different ways, such as safety in expectation, safety in values, safe trajectories, and safe states and actions.
We can formulate this as
\begin{equation}
    \max_{\pi \in \Pi} \mathbb{E}_\pi (G) \,\,\, s.t. \,\,\, c_i= \{ h_i \leq \alpha \},
\end{equation}
\noindent
where $h_i$ can be a function related to the returns, trajectories, values, states, and actions, and $\alpha$ is a safety threshold. 
Consequently, side information can be used in the formulation of such constraints.

On the other hand, Safe Exploration modifies the exploration process subject to external knowledge, which in our case translates to incorporating side information into the exploration process.
While intuitively, this overlaps with the usage of side information for directed exploration, a distinguishing feature of this work is the final goal of this directed exploration, which is to be safe, which might come at the cost of sample efficiency and/or generalization.
\section{Structure as Side Information}
\label{sec:Structure}

Structure can be considered a particular kind of side information about decomposability.
In this section, we discuss the nature of decomposability and the various ways it can be represented. 
In \cref{sec:Structure:types}, we explain the impact of structural side information by explaining how it decomposes complex systems and categorizes such decompositions into four archetypes.
In \cref{sec:Structure:Latent} - \cref{sec:Structure:Modular}, we discuss these archetypes further to connect them with existing literature.

% % Different types of structures
\subsection{Decomposability and Structural Archetypes}
\label{sec:Structure:types}

Decomposability is the property of a system that allows breaking it down into smaller components or subsystems that can be analyzed, understood, and potentially learned more efficiently than the larger system, independently~\shortcite{hofer-book17a}. 
In a decomposable system, the short-term behavior of each subsystem is approximately independent of the short-term behavior of the other subsystems. 
In the long run, the behavior of any subsystem depends on the behavior of the other subsystems only in an aggregated way.

\begin{figure}
    \centering
    \includegraphics[width=\textwidth]{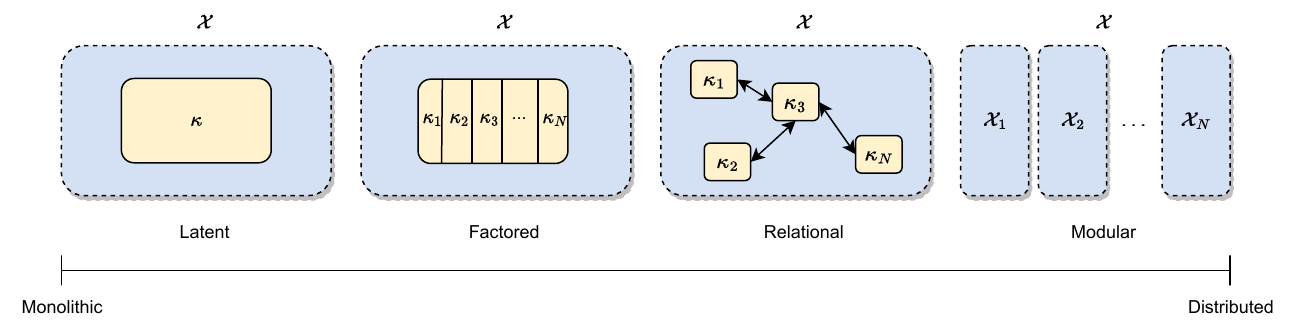}
    \caption{\textbf{Spectrum of Decomposability and Structural Archetypes.} On the left end of the spectrum exist monolithic structural decompositions where a \emph{latent} representation of $\mathcal{X}$ can be learned and incorporated as an inductive bias. Moving towards the right, we can learn multiple latent representations, albeit in a monolithic solution. These are \emph{factored} representations. Further ahead, we see the emergence of interactionally complex decompositions, where knowledge about factorization and how they relate to each other might be essential and can be incorporated using \emph{relational} representations. Finally, we see fully distributed subsystems that can be learned using individual policies. We call these \emph{modular} representations.}
    \label{fig:decomposability}
\end{figure}

Concerning the RL pipeline in \cref{fig:rl-pipeline}, we can see decomposability along two axes: 
\begin{inparaenum}[(i)]
    \item \emph{Problem Decomposition} i.e., the environment parameterization, states, actions, transitions, and rewards; 
    \item \emph{Solution Decomposition} i.e., the learned policies, value functions, models, and training procedures.
    % \item \emph{Training Regime Decomposition} i.e., task decomposition into subtasks and their sequence.
\end{inparaenum}
The spectrum of decomposability~\shortcite{hofer-book17a} provides an intuitive way to understand where a system lies in this regard.
On one end of the spectrum, problems are non-decomposable, while on the other end, problems can be decomposed into weakly interacting sub-problems. 
Similarly, solutions on the former are monolithic, while those on the latter are modular. 
We capture this problem-solution interplay by marking four different archetypes of decomposability.
Decomposition can be incorporated by learning appropriate representations at the granularity of the decomposition, as shown in \cref{fig:decomposability}.
Thus, the following sections use the terms decompositions and representations interchangeably since the representations that we are referring to in this work particularly target decompositions.

\subsection{Latent Decomposition} 
\label{sec:Structure:Latent}
Latent representations can be helpful in complex environments where the underlying structure is unclear or non-stationary.
Under this view, a pipeline component $\mathcal{X}$ can be approximated by a latent representation $\kappa$, which can then be integrated into the learning process. 
The quantities in the learning process that depend on $\mathcal{X}$ can now be re-conditioned on $\kappa$:
    \begin{equation}\label{eq:latent-assumption}
         Z: \mathcal{X} \to \kappa.
    \end{equation}

\paragraph{Latent States and Actions.}
% Latent state spaces have been classically explored through concepts such as Latent MDPs~\shortcite{kwon-neurips21a}, aiming to discover a latent representation of the state space sufficient to learn an optimal policy. 
Latent representations of states are used for tackling scenarios such as rich observation spaces~\shortcite{du-icml19a} and contextual settings~\shortcite{hallak-arxiv15}. 
Latent actions have been similarly explored in settings with stochastic action sets~\shortcite{boutilier-ijcai18a}.  

% Given an encoder $\phi: \mathcal{S} \times \mathcal{A} \to \mathcal{\kappa}$ and a decoder $\beta: \mathcal{\kappa} \to \mathcal{S}$, the latent state-action formulation allows decomposing transition dynamics into a low-rank approximation, 
% \[
% P(s' \mid s,a) = \langle \phi(\kappa \mid s,a)\beta(s' \mid \kappa) \rangle.
% \]

\paragraph{Latent Transition and Rewards.}
While latent states allow decomposing transition matrices, another way to approach the problem directly is to decompose transition matrices into low-rank approximations.
% This can be mathematically shown as
% \[
%     P(s'\mid s,a)  = \phi(s' \mid s,a) \beta(s'\mid s,a).
% \]
Linear MDPs~\shortcite{papini-neurips21} and applications in Model-based RL~\shortcite{vanderpol-aamas20a} have studied this form of direct decomposition. 
A similar decomposition can also be applied to rewards  by assuming the reward signal to be generated from a latent function that can be learned as an auxiliary learning objective~\shortcite{wang-aaai20a}. 

\subsection{Factored Decomposition}
\label{sec:Structure:Factored}
The factored decomposition moves slightly away from the monolithic nature by representing $\mathcal{X}$ using (latent) factors $\mathcal{K} = \{\kappa_1, \dots, \kappa_N$ \}. 
% We assume that the $\mathcal{X}$ can be characterized by a set of factors
A crucial aspect of factorization is that the factors can potentially impose some form of conditional independence in their effects on the learning dynamics.
    \begin{equation}\label{eq:factored-assumption}
        Z: \mathcal{X} \to \mathcal{K}
    \end{equation}
\paragraph{Factored States and Actions.}
Factored state and action spaces have been explored in the Factored MDPs~\shortcite{kearns-ijcai99,boutilier-ai00,guestrin-jair03a}. 
Methods in this setting traditionally capture subsequent state distribution using mechanisms such as Dynamic Bayesian Networks~\shortcite{mihajlovic-misc01a}. 
% Crucially, factored states can lead to conditional independence in the transition dynamics.

% \begin{equation}
%     P(s' \mid s,a)  = \Pi_i P(z_i' \mid s,a) 
% \end{equation}

Factored action representations have also been used for tackling high-dimensional actions~\shortcite{mahajan-arxiv21a}. 
These methods either impose a factorized structure on subsets of a high-dimensional action set~\shortcite{kim-pricai02} or impose this structure through the Q-values that lead to the final action~\shortcite{tang-darl22a}. 
Crucially, these methods can potentially exploit some form of independence resulting from such factorization, either in the state representations or transitions. 

\paragraph{Factored Rewards.}
Combined with factored states or modeled independently, factored rewards have been used to model perturbed rewards~\shortcite{wang-aaai20a}, or to factor latent variable models using causal priors~\shortcite{perez-aaai20a}. 
% A parallel setting of factored rewards has been explored in conjunction with factored states or as independent models. 
% % The fundamental assumption imposed is that rewards can be factored like states according to \cref{eq:factored-assumption}. 
% This factored reward vector $\bm{R} \in \mathbb{R}^N$ can then be used for scenarios such as modeling perturbed rewards in multi-objective settings~\shortcite{mambelli-arxiv22a}.
While Factored MDPs do not naturally lead to factored policies, combining state and reward factorization can lead to factorization of value functions~\shortcite{koller-ijcai99,sodhani-iclr22wsa}.

% \begin{equation}
%     Q^\pi (s,a) = \sum_{i=1}^N Q^\pi (\kappa_i, a)= \sum_{i=1}^N r(\kappa_i,a) + \gamma \sum_{i=1}^N P(\kappa_i' \mid s,a) r(\kappa_i, a) 
% \end{equation}

\subsection{Relational Decomposition}
\label{sec:Structure:Relational}

In addition to representing the problem using a set of factors, we can also utilize information about how different factors interact. 
Usually, these relations exist between entities in a scene and are used to formulate learning methods based on inductive logic~\shortcite{dzeroski-ml01}. 
Traditionally, such relations were limited to first-order logic, but the relational structure has also been captured through graphs. 
Mathematically, side information $Z$ takes the original entity $\mathcal{X}$ as an input and maps it to to a function $\phi$ over the set of factors $\mathcal{K} = \{\kappa_i, \dots \kappa_N \}$. Therefore, $Z$ now outputs functions over $\mathcal{K}$.

\begin{equation}\label{eq:relational-assumption}
    Z: \mathcal{X} \to \phi  \big ( \mathcal{K} \big ) 
\end{equation}

Specifically, $\phi$ is a function that describes relations between groups of $m$ entities ($m$ being the order of the relation) in $\mathcal{K}$. 
Thus, the inputs to $\phi$ are $m$-tuples of factors, which it maps to a multiset of symbols $\{ \psi\}^+$ (such as coordinates, distance measures, or logical predicates).
In other words, $\phi$ describes relations between $m$ input entities using symbols $\psi$, and the multiset allows us to more generally describe the case where similar relations can exist between different entities:

\begin{equation}\label{eq:relational-phi-psi}
     \phi: (\mathcal{K})^m \to \{ \psi \}^+.
\end{equation}

Such representations allow talking about generalization over the relationship between entities in $\mathcal{K}$ and different forms of $\phi$ and help us circumvent the dimensionality of enumerative spaces.

\paragraph{Relational States and Actions.}
Classically, relational representations have been used to model state spaces in Relational MDPs \shortcite{dzeroski-ml01,guestrin-ijcai03a} and Object-Oriented MDPs \shortcite{diuk-icml08a}. 
They represent factored state spaces using first-order representations of objects, predicates, and functions to describe a set of ground MDPs.
Such representations can capture interactionally complex structures between entities much more efficiently.
Additionally, permutations of the interactions between the entities can help define new MDPs that differ in their dynamics, thus contributing towards work in generalization.

States can also be more generally represented as graphs \shortcite{janisch-arxiv20a,sharma-uai22a}, or by using symbolic inductive biases \shortcite{garnelo-arxiv16a} fed to a learning module in addition to the original state. 

Action relations help tackle instances where the agent has multiple possible actions available and the set of actions is significantly large. 
These methods capture relations using either attention mechanisms \shortcite{jain-icml21a,biza-iclr22a} or graphs \shortcite{wang-ijcai19a}, thus offering scalability to high-dimensional action spaces.
Additionally, relations between states and actions have helped define notions such as intents and affordances~\shortcite{abel-icaps15a,khetapal-icml20a}.

\paragraph{Relational Value Functions and Policies.} 
Traditional work in Relational MDPs has also explored ways to represent and construct first-order representations of value functions and/or policies to generalize to new instances. 
These include Regression Trees \shortcite{mausam-icaps03a}, Decision Lists \shortcite{fern-jair06a}, Algebraic Decision Diagrams \shortcite{joshi-jair11a}, Linear Basis Functions \shortcite{guestrin-ijcai03a,sanner-uai05a}, and Graph Laplacians \shortcite{mahadevan-jmlr07a}. 
Recent approaches have started looking into DNN representations~\shortcite{zambaldi-iclr19a,garg-icml20a}, with extensions into modeling problem aspects such as morphology in Robotic tasks~\shortcite{wang-iclr18a} in a relational manner, or learning diffusion operators for constructing intrinsic rewards~\shortcite{klissarov-icml23a}. 

\paragraph{Relational Tasks.}
A parallel line of work looks at capturing relations in a multi-task setting, where task perturbations are either in the form of goals and corresponding rewards \shortcite{sohn-neurips18a,illanes-icaps20a,kumar-neurips22a}. 
Most work aims at integrating these relationships into the optimization procedure and/or additionally capturing them as models. 
We delve deeper into specifics in later sections.

\subsection{Modular Decomposition}
\label{sec:Structure:Modular}

Modular decompositions exist at the other end of the spectrum of decomposability, where individual value functions and/or policies can be learned for each decomposed entity $\mathcal{X}$.
Specifically, a task can be broken down into subsystems $\mathcal{X}_1, \dots, \mathcal{X}_N$ for which models, value functions, and policies can be independently learned. 
\begin{equation}\label{eq:modular-assumption}
    Z: \mathcal{X} \to \{ \mathcal{X}_1, \dots, \mathcal{X}_N \}.
\end{equation}
Such modularity can exist along the following axes:
\begin{inparaenum}[(i)]
    \item \emph{Spatial Modularity} allows learning quantities specific to parts of the state space, thus effectively reducing the dimensionality of the states;
    \item \emph{Temporal Modularity} allows breaking down tasks into sequences over a learning horizon and, thus, learning modular quantities in a sequence;
    \item \emph{Functional Modularity} allows decomposing the policy architecture into functionally modular parts, even if the problem is spatially and temporally monolithic.
\end{inparaenum}

A potential consequence of such breakdown is the emergence of a hierarchy, and when learning problems exploit this hierarchical relationship, these problems come under the purview of Hierarchical RL (HRL)~\shortcite{pateria-acmcs22a}. 
The learned policies can also exhibit a hierarchy, where each can choose the lower-level policies to execute the subtasks. 
Each level can be treated as a planning problem~\shortcite{yang-ijcai18a} or a learning problem~\shortcite{sohn-neurips18a}, thus allowing solutions to combine planning and learning through the hierarchy. 
Hierarchy, however, is not a necessity for modularity. 

\paragraph{Modularity in States and Goals}
Modular decomposition of state spaces is primarily studied at high-level planning and state-abstractions for HRL methods \shortcite{kokel-icaps21a}. 
Approaches such as Q-decomposition~\shortcite{russel-icml03a,bouton-aamas19a} have explored agent design by communicating Q values learned by individual agents on parts of the state-action space to an arbitrator that suggests the following action.
Additionally, the literature on skills has looked into the direction of training policies for individual parts of the state-space \shortcite{goyal-iclr20a}.
Similarly, partial models only make predictions or specific parts of the observation-action spaces in Model-Based settings~\shortcite{talvitie-neurips08,khetarpal-neurips21a}.
Goals have been considered explicitly in methods that either use goals as an interface between levels of hierarchy \shortcite{Kulkarni-neurips16,nachum-neurips18a,gehring-neurips21a}, or as outputs of task specification methods~\shortcite{jiang-neurips19a,illanes-icaps20a}.

\paragraph{Modularity in Actions}
Modularity in action spaces refers to conditioning policies on learned action abstractions. 
The classic example of such methods belongs to the options framework where policies are associated with temporal abstractions over actions~\shortcite{sutton-ai99a}. 
In HRL methods, learning and planning of the higher levels are based on the lower-level policies and termination conditions of their execution.

\paragraph{Compositional Policies}
Continual settings utilize policies compositionally by treating already learned policies as primitives. 
Such methods feed these primitives to the discrete optimization problems for selection mechanisms or to continuous optimization settings involving ensembling~\shortcite{song-arxiv23a} and distillation~\shortcite{rusu-iclr16a}. 
Modularity in such settings manifests itself by construction and is a central factor in building solutions. 
Even though the final policy in such paradigms, obtained through ensembling, selection, and/or distillation, can be monolithic, obtaining such policies is a purely distributed regime.

\section{Patterns of Incorporating Structure}
\label{sec:patterns}
% \todo{format the tables after adding all other content}
Having defined different forms of decomposability and the different objectives that side information can be used to accomplish, we now connect the two by understanding the ways of incorporating structure into a learning process. 
We assume that some form of structure exists in the problem and/or the solution space, which can be incorporated into the learning pipeline as an inductive bias.
To understand how decomposability can be incorporated into the RL pipeline, we could start a potential categorization along two axes: the type of decomposition (Latent, Factored, Relational, and Modular) and the part of the pipeline to which this decomposition is applied (such as states, or actions). 
However, such a categorization ignores a significant part of the process: the method by which the pipeline is conditioned on side information.
For example, information about goals can be used to learn state abstractions or directly fed as input to the policy network.
Both of these design decisions can have very different impacts in practice.
Thus, in addition to the two axes mentioned above, we survey the literature with another specific question:
\emph{Do existing methods incorporate structural information using repeatable design decisions?} 
The answer to this question, inspired by the categorization proposed by \shortciteA{jonschkowski-arxiv15}, brings us to \emph{patterns of incorporating structure}.

% \todo{refine overarching narrative of this section for individual patterns.}

\begin{figure}[ht]
    % \tiny
    \captionsetup[subfigure]{position=b}
    \centering
    % \begin{subfigure}[t]{\textwidth}
            % \centering
             \subcaptionbox{Abstraction\label{fig:abstraction}}{
                 \includegraphics[width=0.23\textwidth]{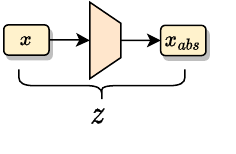}
             }
             \subcaptionbox{Augmentation\label{fig:augmentation}}{
                \includegraphics[width=0.18\textwidth]{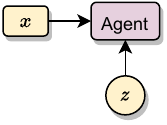}
             }
             \subcaptionbox{Aux. Optimization\label{fig:aux-opt}}{
                \includegraphics[width=0.22\textwidth]{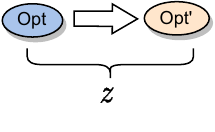}
             }
             \subcaptionbox{Aux. Model\label{fig:aux-model}}{
                \includegraphics[width=0.17\textwidth]{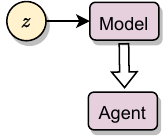}
             }
             \subcaptionbox{Portfolio\label{fig:portfolio}}{
                 \includegraphics[width=0.3\textwidth]{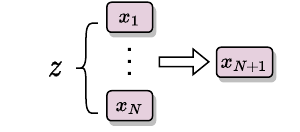}
             }
             \subcaptionbox{Environment Gen.\label{fig:env-gen}}{
                \includegraphics[width=0.24\textwidth]{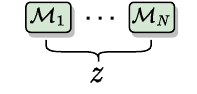}
             }
             \subcaptionbox{Explicitly Designed\label{fig:exp-design}}{
                \includegraphics[width=0.25 \textwidth]{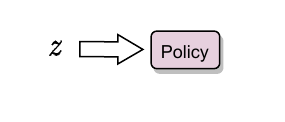}
             }
     \caption{\textbf{Patterns of incorporating structural information.} We categorize the methods of incorporating structure as inductive biases into the learning pipeline into patterns that can be applied for different kinds of usages. Each pattern is shown as a plug-and-play modification of the RL pipeline $\Omega$ that aims to improve the performance of $\Omega$ on one or more objectives discussed in \cref{sec:usage}. 
     }
      \label{fig:patterns}  
\end{figure}

A pattern is a principled change in the RL pipeline $\Omega$ that allows the pipeline to achieve one, or a combination of, the additional objectives: \emph{Sample Efficiency}, \emph{Generalization}, \emph{Safety}, and \emph{Interpretability}.
We categorize the literature into seven patterns, an overview of which has been shown in \cref{fig:patterns}.
% Each pattern

\begin{figure}
    \begin{center}
        \includegraphics[width=0.7\textwidth, trim={4.2cm 8cm 4.4cm 2cm}]{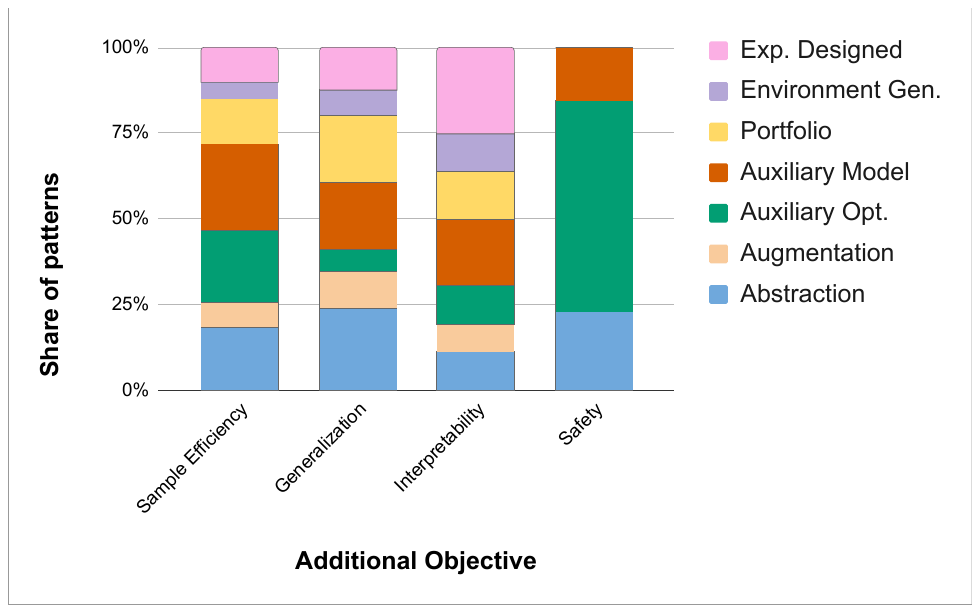}
        % stable= new-plot-3
        \caption{\textbf{Proclivities.} A meta-analysis of the proclivities of each pattern to the additional objectives. The four additional objectives covered in this text are on the x-axis. We show each objective's share percentage of publications utilizing individual patterns. This data has been shown on the y-axis with different colors for each pattern. Therefore, this figure helps us understand correlations between patterns and the kind of objectives they address.}
    \end{center}
    % \begin{minipage}[c]{0.5\textwidth}
        % \includegraphics[width=\textwidth,trim={4.2cm 16cm 4.4cm 3cm},clip]{figures/pattern-proclivities-new.pdf}
    
    % \end{minipage}\hfill
    % \begin{minipage}[c]{0.5\textwidth}
    
    \label{fig:pattern-proclivities}
    % \end{minipage}
\end{figure}

Our proposed patterns come from our literature survey and are meant to provide an initial direction for such categorization. 
We do not consider this list exhaustive but more as a starting point upon which to build further. 
% While exploration is a central piece of the RL puzzle, our survey shows that most of the prior works that use the structure for directed exploration can be subsumed into the four objectives previously mentioned. 
We present an overview of our meta-analysis on the patterns used for which of the four use cases in \Cref{fig:pattern-proclivities}.

% \begin{wrapfigure}{r}{0.35\textwidth}
%     \centering
%     \includegraphics[width=0.5\textwidth]{figures/pattern-proclivities.pdf}
%     \vspace{-20pt}
    
% \end{wrapfigure} 

% \begin{figure}
%     \centering
%     \includegraphics[width=0.7\textwidth]{figures/pattern-proclivities.pdf}
%     \caption{\textbf{Proclivities.} A meta-analysis of the proclivities of each pattern to the additional objectives. On the x-axis are the patterns discussed in this text. For each of these patterns, we analyze each publication for the kind of additional objectives that it addresses.
%     Using this data, we understand the proclivity of each pattern to each of hte four additional objectives by calculating the percentage of methods that use the particular pattern to address each objective. This data has been shown on the y-axis
%     % while on the y-axis are the percentage of publications for each additional objective that addresses it using a particular pattern.
%     }
%     \label{fig:pattern-proclivities}
% \end{figure}

% \begin{figure}[!ht]
%     % \tiny
%     \captionsetup[subfigure]{position=b}
%     \centering
%     % \begin{subfigure}[t]{\textwidth}
%             % \centering
%              \subcaptionbox{Patterns\label{fig:pattern-proclivities}}{
%                  \includegraphics[width=0.4\textwidth]{figures/pattern-proclivities.pdf}
%              }
%              \subcaptionbox{Decompositions\label{fig:decomposition-proclivities}}{
%                 \includegraphics[width=0.4\textwidth]{figures/decomposition-proclivities.pdf}
%                 \vspace{20pt}
%              }
%      \caption{A meta-analysis of the proclivities of each pattern and decomposition to auxiliary objectives. \todo{align figures}}  
% \end{figure}

In the following subsections, we delve deeper into each pattern, explaining different lines of literature that apply each pattern for different use cases.
To further provide intuition about this categorization, we will consider the running example of a taxi service, where the task of the RL agent (the taxi) is to pick up passengers from various locations and drop them at their desired destinations within a city grid. 
The agent receives a positive reward when a passenger is successfully dropped off at their destination and incurs a minor penalty for each time step to encourage efficiency.

For each of the following sections, we present a table of the surveyed methods that categorizes the work in the following manner: 
\begin{inparaenum}[(i)]
    \item The structured space, information about which is incorporated as side information;
    \item The type of decomposition exhibited for that structured space. We specifically categorize works that use structured task distributions through goals and/or rewards;
    \item The additional objectives for which the decomposition is utilized. 
    % Thus, we address these individually for each pattern. 
\end{inparaenum}
Our rationale behind the table format is to highlight the areas where further research might be lucrative in addition to categorizing the existing literature. 
These are the spots in the tables where we could not yet find literature, and we believe additional work can be important; therefore, in addition to categorizing existing methods, empty areas in the table highlight avenues for future research.
% Therefore, in addition to categorizing existing methods, empty areas in the table 

\subsection{Abstraction Pattern} 
\label{sec:patterns:abstraction}

Abstraction pattern utilizes structural information to create abstract entities in the RL pipeline.
For any entity, $X$, an abstraction utilizes the structural information to create $X_{abs}$, which takes over the role of $X$ in the learning procedure.
% Example
In the taxi example, the state space can be abstracted to the taxi's current grid cell, the destination grid cell of the current passenger, and whether the taxi is currently carrying a passenger. 
This significantly simplifies the state space compared to representing the full details of the city grid. 
The action space could also be abstracted to moving in the four cardinal directions, plus picking up and dropping off a passenger.
Behavioral abstractions are intricately related to history-based abstractions in that they address similar applications, namely, abstraction states and histories, into lower dimensional representations that can be leveraged for RL. 
Therefore, we address both of these in the following sections.
% Important consideration
% \begin{wrapfigure}{r}{0.35\textwidth}
%     \centering
%     \vspace{10pt}
%     \includegraphics[width=0.3\textwidth, trim={0cm 0.4cm 0.4cm 0cm}]{figures/patterns/Abstraction.pdf}
%     \caption{Abstraction pattern}
%     % \vspace{-10pt}
% \end{wrapfigure} 

\begin{figure}[htb]
    \centering
    \fbox{\includegraphics[width=0.3\textwidth, trim={0cm 0.4cm 0.4cm 0cm}]{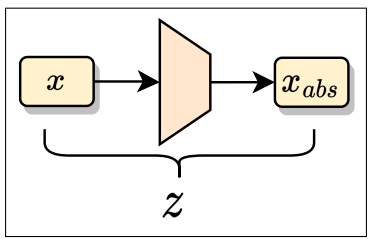}}
    \caption{Abstraction Pattern}
\end{figure} 

Finding appropriate abstractions can be a challenging task in itself.
Too much abstraction can lead to loss of critical information, while too little might not significantly reduce complexity~\shortcite{dockhorn-book23a}.
Consequently, learning-based methods that jointly learn abstractions factor this granularity into the learning process.

Abstractions have been thoroughly explored in the literature, with early work addressing a formal theory on state abstractions~\shortcite{li-aim06a,sutton-book18b}.
Recent works have primarily used abstractions for tackling generalization. 
Thus, we see in \cref{fig:pattern-proclivities} that generalization is the most explored use case for abstractions.
However, the advantages mentioned earlier of abstraction usually interleave these approaches with sample efficiency gains and safety. 
Given the widespread use of abstractions in the literature, the following paragraphs explore how different abstractions impact each use case.

\begin{center}
    \scriptsize
    % \centering
    
    % \caption{Overview of Literature using the Abstraction pattern}
    \begin{tabularx}{\textwidth}{Y|Y|X|X|X|X}
    
    \toprule[1pt]\midrule[0.3pt]
    % \rowcolor{LightCyan}
    \textbf{Space}          & \textbf{Type}    & \multicolumn{1}{c}{\textbf{Efficiency}}   & \multicolumn{1}{c}{\textbf{Generalization}}   & \multicolumn{1}{c}{\textbf{Interpretabiltiy}} & \multicolumn{1}{c}{\textbf{Safety}} \\  
    \hline
    % \toprule[1pt]
    \multirow{3}{*}{Goals}    & Latent                    & \shortciteA{gallouedec-icml23a} &\shortciteA{estruch-icml22}, \shortciteA{gallouedec-icml23a} & &\\ 
                            \cline{2-6}
                            & Relational                & & & \shortciteA{prakash-arxiv22a} & \\
                            \cline{2-6}
                            & Modular                   & \shortciteA{icarte-jair22a} & \shortciteA{icarte-jair22a} & \shortciteA{prakash-arxiv22a}, \shortciteA{icarte-jair22a} &\\
                            
    \hline
    % \midrule[0.3pt]
    \multirow{4}*{States}   & Latent                    & \shortciteA{zhang-iclr23a}, \shortciteA{ghorbani-arxiv20a}, \shortciteA{allen-neurips21a}, \shortciteA{zhang-iclr21c}, \shortciteA{gelada-icml19a}, \shortciteA{lee-neurips20a}, \shortciteA{azizzadenesheli-arxiv16a}, \shortciteA{misra-icml20a} & \shortciteA{gelada-icml19a}, \shortciteA{zhang-icml20a}, \shortciteA{misra-icml20a}, \shortciteA{lee-neurips20a}, \shortciteA{zhang-iclr21b}, \shortciteA{agarwal-iclr21a} & \shortciteA{gillen-corl21a} &\shortciteA{yang-aaai22a}, \shortciteA{gillen-corl21a}  \\
                            \cline{2-6}
                            & Factored                  & \shortciteA{sodhani-iclr22wsa} &\shortciteA{higgins-icml17a}, \shortciteA{perez-aaai20a}, \shortciteA{sodhani-icml21a}, \shortciteA{sodhani-iclr22wsa}, \shortciteA{dunion-iclr23a}, \shortciteA{dunion-neurips23a} & \shortciteA{sodhani-icml21a}, \shortciteA{bewley-aamas22a}, \shortciteA{kooi-arxiv22a} & \\
                            \cline{2-6}
                            & Relational                &  \shortciteA{martinez-ai17a}, \shortciteA{garnelo-arxiv16a}, \shortciteA{kipf-iclr20a}, \shortciteA{kokel-icaps21a}, \shortciteA{klissarov-icml23a} & \shortciteA{janisch-arxiv20a}, \shortciteA{kokel-icaps21a}, \shortciteA{bapst-icml19a}, \shortciteA{adjodah-neurips18a}, \shortciteA{garnelo-arxiv16a}, \shortciteA{kipf-iclr20a}, \shortciteA{karia-ijcai22a} & \shortciteA{adjodah-neurips18a}, \shortciteA{garnelo-arxiv16a} & \\
                            \cline{2-6}
                            & Modular                   & \shortciteA{kokel-icaps21a}, \shortciteA{icarte-jair22a}, \shortciteA{blanco-jair21a} & \shortciteA{kokel-icaps21a}, \shortciteA{steccanella-arxiv21a}, \shortciteA{icarte-jair22a}, \shortciteA{blanco-jair21a} & \shortciteA{icarte-jair22a}, \shortciteA{blanco-jair21a} &\\
    \hline
    % \midrule[0.3pt]
    \multirow{4}*{Actions}  & Latent                    & \shortciteA{zhao-naac19a}, \shortciteA{chandak-icml19a} & & &\\
                            \cline{2-6}
                            & Factored                  & & \shortciteA{perez-aaai20a} & \shortciteA{bewley-aamas22a} & \\
                            \cline{2-6}
                            & Relational                & \shortciteA{christodoulou-arxiv19a} &\shortciteA{bapst-icml19a}  & &\\
                            \cline{2-6}
                            & Modular                   & \shortciteA{blanco-jair21a} & \shortciteA{steccanella-arxiv21a}, \shortciteA{blanco-jair21a} & \shortciteA{blanco-jair21a} &\\
                                        
    \hline
    % \midrule[0.3pt]
    \multirow{2}*{Rewards}  & Latent                    & & \shortciteA{zhang-iclr21b}, \shortciteA{barreto-neurips17a}, \shortciteA{barreto-icml18a}, \shortciteA{borsa-arxiv16a} & & \\
                            \cline{2-6}
                            & Factored                  & \shortciteA{sodhani-iclr22wsa} & \shortciteA{perez-aaai20a}, \shortciteA{sodhani-iclr22wsa}, \shortciteA{sodhani-icml21a},  & \shortciteA{sodhani-icml21a}& \shortciteA{wang-aaai20a}\\                
    \hline
    % \midrule[0.3pt]
    \multirow{3}*{Dynamics} & Latent                    & \shortciteA{zhang-arxiv20b} & \shortciteA{zhang-arxiv20b}, \shortciteA{borsa-iclr19a}, \shortciteA{perez-aaai20a}, \shortciteA{zhang-iclr21b} & &\\
                            \cline{2-6}
                            & Factored                  & \shortciteA{fu-icml21a} & \shortciteA{fu-icml21a}  & &\\
                            \cline{2-6}
                            & Modular                   & \shortciteA{sun-aaai21a} & \shortciteA{sun-aaai21a}  & &\\
                                         
    \midrule[0.3pt]\bottomrule[1pt]
    \caption{Survey of literature utilizing the abstraction pattern}
    \end{tabularx}
    \label{table:patterns:abstraction}
\end{center}

\paragraph{Generalization.}
%States
State abstractions are a standard choice for improving generalization performance by capturing shared dynamics across MDPs into abstract state spaces using methods such as Invariant Causal Prediction~\shortcite{peters-meth15a,zhang-icml20a}, similarity metrics~\shortcite{zhang-iclr21b,estruch-icml22,castro-neurips21a,lan-aaai21a,agarwal-iclr21a,lan-iclr23a,castro-tmlr23a}, Free Energy Minimization~\shortcite{ghorbani-arxiv20a}, and disentanglement~\shortcite{higgins-icml17a,burgess-arxiv19a,kooi-arxiv22a,dunion-iclr23a,dunion-neurips23a}.

% Q values
Value functions serve as temporal abstractions for shared dynamics in Multi-task Settings. 
Successor Features (SF)~\shortcite{dayan-nc93a,barreto-neurips17a} exploit latent reward and dynamic decompositions using value functions as an abstraction. 
Subsequent works have combined them with Generalized Policy Iteration~\shortcite{barreto-icml18a} and Universal Value Function Approximators~\shortcite{schaul-icml15a,borsa-iclr19a}.
Factorization in value functions, on the other hand, helps improve sample efficiency and generalization both~\shortcite{sodhani-icml21a,sodhani-iclr22wsa}.
 
%Relational
Relational abstractions contribute to generalization by incorporating symbolic spaces into the RL pipeline.
These help incorporate planning approaches in hierarchical frameworks~\shortcite{janisch-arxiv20a,kokel-icaps21a}. 
Additionally, relational abstractions can help abstract away general aspects of a collection of MDPs, thus allowing methods to learn generalizable Q-values over abstract states and actions that can be transferred to new tasks~\shortcite{karia-ijcai22a} or develop methods specifically for graph-structured spaces~\shortcite{bapst-icml19a,kipf-iclr20a}.

%Modular
Abstractions can additionally enable generalization in hierarchical settings by compressing state spaces~\shortcite{steccanella-arxiv21a}, abstract automata~\shortcite{blanco-jair21a,icarte-jair22a}, templates of dynamics across tasks~\shortcite{sun-aaai21a}, or even be combined with options to preserve optimal values~\shortcite{abel-aistats20a}.

\paragraph{Sample Efficiency.}

% Latent states
Latent variable models improve sample efficiency across the RL pipeline.
Latent state abstractions help improve sample efficiency in Model-based RL~\shortcite{gelada-icml19a} and also help improve the tractability of policy learning over options in HRL~\shortcite{steccanella-arxiv21a}.
In model-free tasks, these are also learned as inverse models for visual features~\shortcite{allen-neurips21a} or control in a latent space~\shortcite{lee-neurips20a}.
Latent transition models demonstrate efficiency gains by capturing task-relevant information in noisy settings~\shortcite{fu-icml21a}, by preserving bisimulation distances between original states~\shortcite{zhang-iclr21c}, or by utilizing factorized abstractions~\shortcite{perez-aaai20a}.
Learned latent abstractions~\shortcite{gallouedec-icml23a} also contribute to the exploration mechanism in the Go-Explore regime~\shortcite{ecoffet-nature21a}.
% Additionally, Latent variable models~\shortcite{perez-aaai20a} utilize factorization as abstractions to impose independence conditions in the transition dynamics.

%Latent actions
Latent action models expedite convergence of policy gradient methods such as REINFORCE~\shortcite{williams-ML92} by shortening the learning horizon in stochastic scenarios like dialog generation~\shortcite{zhao-naac19a}.
Action embeddings, on the other hand, help reduce the dimensionality of large action spaces~\shortcite{chandak-icml19a}
% learn an embedding space of actions using Supervised Learning and then train a policy on this latent space instead of the full action space for model-based RL.

\paragraph{Safety and Interpretability.}
Relational abstractions are an excellent choice for interpretability since they capture interactionally complex decompositions. 
The combination of object-centric representations and learned abstractions adds transparency~\shortcite{adjodah-neurips18a} while symbolic interjections, such as tracking the relational distance between objects, help improve performance ~\shortcite{garnelo-arxiv16a}.

State and reward abstractions help with safety.
Latent states help to learn safe causal inference models by embedding confounders~\shortcite{yang-aaai22a}.
On the other hand, meshes \shortcite{talele-arxiv19a,gillen-corl21a} help benchmark metrics such as robustness in a learned policy.

% \paragraph{Exploration}
% Using abstractions for directed exploration shared overlap with generalization and sample efficiency use cases. 
% \shortciteA{ghorbani-arxiv20a}, on the other hand, promotes explicit exploration by minimizing free energy to learn approximate state abstractions. 
% The usage of Thompson sampling and behavioral policy for each subspace allows them to tackle the exploration problem better.

\subsection{Augmentation Pattern}
\label{sec:patterns:augmentation}

The augmentation pattern treats $X$ and $z$ as separate input entities for the action-selection mechanism. 
The combination can range from the simple concatenation of structural information to the state or actions to more involved methods of conditioning policy or value functions on additional information.
Crucially, the structural information neither directly influences the optimization procedure nor changes the nature of $X$. 
It simply augments the already existing entities.
In this view, abstractions learned in an auxiliary manner and concatenated to states, actions, or models can also be considered augmentations since the original entity remains unchanged.

% \begin{wrapfigure}{r}{0.35\textwidth}
%     \centering
%     \vspace{-10pt}
%     \framebox[0.35\textwidth]{\includegraphics[width=0.25\textwidth, trim={0.3cm 0.1cm 0.2cm 0cm}]{figures/patterns/Augmentation.pdf}}
%     \vspace{-20pt}
% \end{wrapfigure} 

\begin{figure}[htb]
    \centering
    % \vspace{-10pt}
    \framebox[0.3\textwidth]{\includegraphics[width=0.25\textwidth, trim={0.2cm 0.1cm 0.2cm 0cm}]{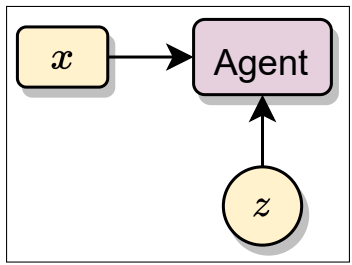}}
    % \vspace{-20pt}
    \caption{Augmentation Pattern}
\end{figure}

% Example
For the taxi example, one way to apply the augmentation pattern would be by conditioning the policy on additional information, such as the time of day or day of the week. 
This information could be helpful because traffic conditions and passenger demands can vary depending on these factors.
% Downside
However, augmentations can increase the complexity of the policy and the learning process, and care needs to be taken to ensure that the policy does not overfit the additional information.
Consequently, this pattern is generally not explored as much as abstraction. 
While we see usages of augmentations equitably for most use cases in \cref{fig:pattern-proclivities}, the number of papers utilizing this pattern still falls short compared to more established techniques, such as abstraction. 
In the following paragraphs, we delineate three augmentations in the surveyed work.

\paragraph{Context-based Augmentations.}
Contextual representations of dynamics~\shortcite{sodhani-ldcc22a,guo-iclr22a} and goal-related information~\shortcite{nachum-neurips18a,islam-neurips22a}  help with generalization and sample efficiency by exposing the agent to the necessary information for optimality.
Goal augmentations additionally allow interpretable mechanisms for specifying goals~\shortcite{beyret-iros19a}.
On the other hand, augmentation of meta-learned latent spaces to the normal state can promote temporally coherent exploration across tasks~\shortcite{gupta-neurips18a}.
Action histories~\shortcite{tennenholtz-icml19a} can directly help with sample efficiency, and action relations~\shortcite{jain-icml20a,jain-icml21a} contribute to generalization over large action sets.

% \begin{table}[htbp]
\begin{center}
    \scriptsize
    \centering
    % \caption{Overview of Literature using the Augmentation pattern}
     
    \begin{tabularx}{\textwidth}{Y|Y|X|X|X|X}
    \toprule[1pt]\midrule[0.3pt]
    \textbf{Space}              & \textbf{Type}             & \multicolumn{1}{c}{\textbf{Efficiency}}  & \multicolumn{1}{c}{\textbf{Generalization}} & \multicolumn{1}{c}{\textbf{Interpretabiltiy}} & \multicolumn{1}{c}{\textbf{Safety}} \\  
    \hline
    \multirow{4}{*}{Goals}      & Latent                    &  & \shortciteA{andreas-naacl18a}, \shortciteA{schaul-icml15a} & & \\ 
                                \cline{2-6} 
                                & Factored                  & \shortciteA{islam-neurips22a} & \shortciteA{jiang-neurips19a} & & \\ 
                                \cline{2-6}
                                & Relational                & \shortciteA{andreas-naacl18a} & \shortciteA{andreas-naacl18a}, \shortciteA{jiang-neurips19a} & & \\ 
                                \cline{2-6}
                                & Modular                   & \shortciteA{gehring-neurips21a}, \shortciteA{beyret-iros19a} & \shortciteA{jiang-neurips19a}, \shortciteA{gehring-neurips21a} & \shortciteA{beyret-iros19a} & \\ 
    \hline
    % \midrule[0.3pt]
    \multirow{4}{*}{States}     & Latent                    & \shortciteA{islam-neurips22a}, \shortciteA{andreas-naacl18a}, \shortciteA{gupta-neurips18a} & \shortciteA{andreas-naacl18a}, \shortciteA{sodhani-ldcc22a}, \shortciteA{gupta-neurips18a} & & \\
                                \cline{2-6}
                                & Factored                  & \shortciteA{islam-neurips22a} & & & \\ 
                                \cline{2-6}
                                & Relational                & \shortciteA{andreas-naacl18a} & \shortciteA{andreas-naacl18a} & & \\ 
                                \cline{2-6}
                                & Modular                   & & & & \\
    \hline
    % \midrule[0.3pt]
    \multirow{3}{*}{Actions}    & Latent                    & \shortciteA{tennenholtz-icml19a} & \shortciteA{jain-icml21a}, \shortciteA{jain-icml20a} & & \\ 
                                \cline{2-6} 
                                & Relational                & & \shortciteA{jain-icml21a} & & \\ 
                                \cline{2-6}
                                & Modular                   & \shortciteA{devin-neurips19a} & \shortciteA{pathak-neurips19a}, \shortciteA{devin-neurips19a} & & \\
    \hline
    % \midrule[0.3pt]
    Rewards                     & Factored                  & \shortciteA{huang-icml20a} & \shortciteA{huang-icml20a} & & \\  
    \hline
    % \midrule[0.3pt]                             
    \multirow{2}{*}{Dynamics}   & Latent                    & \shortciteA{wang-icml22a} & \shortciteA{sodhani-ldcc22a}, \shortciteA{guo-iclr22a}, \shortciteA{wang-icml22a} & & \\ 
                                \cline{2-6} 
                                & Factored                  & & \shortciteA{goyal-iclr21a} & & \\                                 
    \hline
    % \midrule[0.3pt]
    Policies                    & Modular                   & \shortciteA{raza-icit19a}, \shortciteA{haarnoja-icml18b}, \shortciteA{marzi-arxiv23a} & \shortciteA{haarnoja-icml18b} & \shortciteA{verma-icml18} & \\             
    \midrule[0.3pt]\bottomrule[1pt]
    \caption{Survey of literature utilizing the augmentation pattern}
    \end{tabularx}
    \label{table:patterns:augmentation}
\end{center}
% \end{table}

\paragraph{Language Augmentations.}
Language explicitly captures relational metadata in the world. 
Latent language interpretation models~\shortcite{andreas-naacl18a} utilize the compositionality of language to achieve better exploration and generalization to different relational settings, as represented by their language descriptions.
On the other hand, goal descriptions~\shortcite{jiang-neurips19a} help hierarchical settings by exploiting semantic relationships between different subtasks and producing better goals for lower-level policies. 
Augmentations additionally help make existing methods more interpretable through methods such as the one proposed by \shortciteA{verma-icml18} by guiding search over approximate policies written in human-readable formats.

\paragraph{Control Augmentations.}
Augmentations help with primitive control, such as multi-level control in hierarchical settings. 
Augmenting internal latent variables conditioned on primitive skills~\shortcite{haarnoja-icml18b,gehring-neurips21a,devin-neurips19a} helps tackle sample efficiency in hierarchical settings. 
Augmentations additionally help morphological control~\shortcite{huang-icml20a} by modeling the limbs as individual agents that must learn to join together into a morphology to solve a task~\shortcite{pathak-neurips19a}.

\subsection{Auxiliary Optimization Pattern}
\label{sec:patterns:aux-opt}

This pattern uses structural side information to modify the optimization procedure.
This includes methods involving contrastive losses, reward shaping, concurrent optimization, masking strategies, regularization, baselining, etc.
However, given that the changes in the optimization can go hand-in-hand with modifications of other components, this pattern shares methods with many other patterns. 
For example, contrastive losses can be used to learn state abstractions.
Similarly, a learned model can be utilized for reward shaping as well.
Thus, methods that fall into this category simultaneously utilize both patterns.

% \begin{wrapfigure}{r}{0.35\textwidth}
%     \centering
%     \vspace{-10pt}
%     \framebox[0.35\textwidth]{\includegraphics[width=0.3\textwidth, trim={0.2cm 0.4cm 0.2cm 0cm}]{figures/patterns/Aux-Opt.pdf}}
%     \vspace{-20pt}
% \end{wrapfigure} 

\begin{figure}[htb]
    \centering
    \vspace{-10pt}
    \framebox[0.3\textwidth]{\includegraphics[width=0.25\textwidth, trim={0.2cm 0.4cm 0.2cm 0cm}]{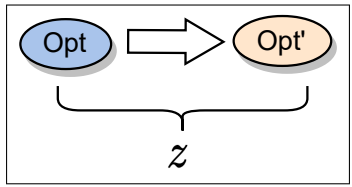}}
    % \vspace{-20pt}
    \caption{Auxiliary Optimization Pattern}
\end{figure}

%Example
In the case of the taxi, reward shaping could help the policy to be reused for slight perturbances in the city grid, where the shaped reward encourages the taxi to stay near areas where passengers are frequently found when it does not have a passenger. 
It is crucial to ensure that the modified optimization process remains aligned with the original objective, i.e., some form of regularization that controls how the modification of the optimization procedure respects the original objective needs to exist.
For reward shaping techniques, this amounts to the invariance of the optimal policy under the shaped reward~\shortcite{ng-icml99a}. 
For auxiliary objectives, this manifests in some form of entropy~\shortcite{fox-uai16a} or divergence regularization~\shortcite{eysenbach-iclr19a}.
Constraints ensure this through recursion~\shortcite{lee-modem22a}, while baselines control the variance of updates~\shortcite{wu-iclr18a}.
The most vigorous use of constraints is in the safety literature, where constraints either help control the updates using some safety criterion or constrain the exploration. 
Consequently, in \cref{fig:pattern-proclivities}, the auxiliary optimization pattern peaks in its proclivity towards addressing safety.
In the following paragraphs, we cover methods that optimize individual aspects of the optimization procedure, namely, rewards, learning objectives, constraints, and parallel optimization.

% \tcbset{colback=gray!5!white,colframe=gray!75!black}
% \begin{tcolorbox}[leftrule=2mm]
% \begin{itemize}
%     \item We see sample efficiency dominating the aux-opt pattern. 
%     \item safe RL uses a significant amount of auxiliary optimization work
%     \item Reward modification in Piedmont
%     \item What are the open areas of work
%     \item Any other interesting point about this pattern? 
% \end{itemize}
% \end{tcolorbox}

% \begin{table}[htbp]
    
%     \label{table:patterns:aux-opt}
\begin{center}
    \scriptsize
    \centering
    % \caption{Overview of Literature using the Auxiliary Optimization pattern}
    \begin{tabularx}{\textwidth}{Y|Y|X|X|X|X}
    \toprule[1pt]\midrule[0.3pt]
    \textbf{Space}              & \textbf{Type}        & \multicolumn{1}{c}{\textbf{Efficiency}}  & \multicolumn{1}{c}{\textbf{Generalization}} & \multicolumn{1}{c}{\textbf{Interpretabiltiy}} & \multicolumn{1}{c}{\textbf{Safety}} \\  
    \hline
    \multirow{3}{*}{Goals}      & Latent                        & & \shortciteA{wang-icml23a} & & \\
                                \cline{2-6}
                                & Relational                    & & \shortciteA{kumar-neurips22a} & & \\ 
                                \cline{2-6}
                                & Factored                      & & & \shortciteA{alabdulkarim-arxiv22a} & \\
                                \cline{2-6}
                                & Modular                       & \shortciteA{nachum-neurips18a}, \shortciteA{illanes-icaps20a}, \shortciteA{li-iclr21c}, \shortciteA{gehring-neurips21a} & & &  \\
    \hline
    % \midrule[0.3pt]
    \multirow{4}{*}{States}     & Latent                        & \shortciteA{mahajan-arxiv17a}, \shortciteA{li-iclr21c}, \shortciteA{azizzadenesheli-arxiv16a}, \shortciteA{ok-neurips18a}, \shortciteA{amin-icml21a}, \shortciteA{nachum-neurips18a}, \shortciteA{ghorbani-arxiv20a}, \shortciteA{yang-iclr20b}, \shortciteA{henaff-neurips22a} & & \shortciteA{harutyunyan-aistats19a} & \shortciteA{zhang-neurips20a}, \shortciteA{yu-icml22a} \\
                                \cline{2-6}
                                & Factored                      & \shortciteA{tavakol-rs14a}, \shortciteA{trimponias-arxiv23a}, \shortciteA{ross-uai08}, \shortciteA{lyu-entropy23a} & & & \shortciteA{lee-modem22a} \\
                                \cline{2-6}
                                & Relational                    & \shortciteA{li-iclr21c} & & & \\
                                \cline{2-6}
                                & Modular                       & \shortciteA{nachum-neurips18a}, \shortciteA{khetarpal-aaai20a} & & \shortciteA{lyu-aaai19} & \\
    \hline
    % \midrule[0.3pt] 
    \multirow{3}{*}{Actions}    & Latent                        & \shortciteA{ok-neurips18a}, \shortciteA{amin-icml21a}, \shortciteA{yang-iclr20b}, \shortciteA{lyu-entropy23a} & \shortciteA{gupta-iclr17a} & \shortciteA{zhang-neurips21ws} & \shortciteA{zhang-axiv19a}, \shortciteA{zhang-arxiv19b}, \shortciteA{zhang-neurips21ws} \\
                                \cline{2-6}
                                & Factored                      & \shortciteA{balaji-neurips20a}, \shortciteA{wu-iclr18a} \shortciteA{metz-arxiv17a}, \shortciteA{spooner-neurips21a}, \shortciteA{tang-darl22a}, \shortciteA{khamassi-irc17a}, \shortciteA{tavakol-rs14a} & & &  \\
                                \cline{2-6}
                                & Modular                       & \shortciteA{metz-arxiv17a}, \shortciteA{klissarov-icml23a} & & \shortciteA{lyu-aaai19} & \shortciteA{jain-ker21a}\\
    \hline
    % \midrule[0.3pt]
    Rewards                     & Factored                      & \shortciteA{trimponias-arxiv23a}, \shortciteA{saxe-icml17a}, \shortciteA{huang-icml20a} & \shortciteA{belogolovsky-ml21a}, \shortciteA{saxe-icml17a},  \shortciteA{buchholz-mmor19a}, \shortciteA{huang-icml20a} & & \shortciteA{prakash-aaai20}, \shortciteA{baheri-arxiv20a} \\
    \hline
    % \midrule[0.3pt]
    \multirow{3}{*}{Dynamics}   & Latent                          & \shortciteA{mu-neurips22a}, \shortciteA{henaff-neurips22a} & \shortciteA{lee-neurips21a}& &  \\
                                \cline{2-6}
                                & Factored                      & \shortciteA{liao-arxiv21a} & \shortciteA{belogolovsky-ml21a}, \shortciteA{buchholz-mmor19a} & & \\
                                \cline{2-6}
                                & Relational                      & \shortciteA{mu-neurips22a}, \shortciteA{illanes-icaps20a} & & & \\
    \hline
    % \midrule[0.3pt]
    Policies                & Latent                        & \shortciteA{hausman-iclr18a} & \shortciteA{hausman-iclr18a}, \shortciteA{gupta-iclr17a} & & \\
    \midrule[0.3pt]\bottomrule[1pt]
    \caption{Survey of literature utilizing the auxiliary optimization pattern}
    \end{tabularx}
    \label{table:patterns:aux-opt}
\end{center}
% \end{table} 

\paragraph{Reward Modification.}
Reward shaping is a common way to incorporate additional information into the optimization procedure.
Methods gain sample efficiency by exploiting modular and relational decompositions through task descriptions~\shortcite{illanes-icaps20a}, or goal information from a higher level policy with off-policy modification to the lower level transitions~\shortcite{nachum-neurips18a}.
Histories of rewards~\shortcite{mahajan-arxiv17a} help learn symmetric relationships between states and, thus, improve the selection procedure for states in a mini-batch for optimization.
Factorization of states and rewards into endogenous and exogenous factors~\shortcite{trimponias-arxiv23a}, on the other hand, helps with safety and sample efficiency through reward corrections.

Extrinsic Rewards can also guide the exploration process. 
Symbolic planners with relational representations interact with a primitive learning policy through extrinsic rewards in hierarchical settings, thus adding interpretability while directly impacting the exploration through the extrinsic reward~\shortcite{lyu-aaai19}.
Alternatively, additional reward sources help determine the quality of counterfactual trajectories, which consequently help explain why an agent took certain kinds of actions~\shortcite{alabdulkarim-arxiv22a}.
Additionally, running averages of rewards help adaptively tune exploration parameters for heterogeneous action spaces~\shortcite{khamassi-irc17a}

On the other hand, intrinsic rewards help explore sparse reward environments.
Latent decompositions help improve such methods by directly impacting the exploration.
Language abstractions serve as latent decompositions used separately for exploration~\shortcite{mu-neurips22a}. 
Alternatively, geometric structures can compare state embeddings and provide episodic bonuses~\shortcite{henaff-neurips22a}.

\paragraph{Auxiliary Learning Objectives.}
Skill-based methods transfer skills between agents with different morphology by learning invariant subspaces and using those to create a transfer auxiliary objective (through a reward signal)~\shortcite{gupta-iclr17a}, or an entropy-based term for policy regularization~\shortcite{hausman-iclr18a}.
Discovering appropriate sub-tasks~\shortcite{solway-ploscb14a} in hierarchical settings is a highly sample-inefficient process. 
\shortciteA{li-iclr21c} tackle this by composing values of the sub-trajectories under the current policy, which they subsequently use for behavior cloning.
Latent decompositions additionally help with robustness and safety when used for some form of policy regularization~\shortcite{zhang-neurips20a}.
Auxiliary losses usually help with generalization and are an excellent entry point for human-like inductive biases~\shortcite{kumar-neurips22a}.
Metrics inspired by the geometry of latent decompositions help learn optimal values in multi-task settings~\shortcite{wang-icml23a}.

\paragraph{Constraints and Baselines.}
Constrained optimization is commonplace in Safe RL, and incorporating structure helps improve the sample efficiency of such methods while making them more interpretable.
Factorization of states into safe and unsafe states helps develop persistent safety conditions~\shortcite{yu-icml22a}, or language abstractions~\shortcite{prakash-aaai20}.
Recursive constraints~\shortcite{lee-modem22a} help explicitly condition the optimization on a latent subset of safe actions using factored states. 
Restricting the exploration of options to non-risky states helps incorporate safety in hierarchical settings as well~\shortcite{jain-ker21a}.
Factorized actions additionally improve the sample efficiency of policy gradient methods through baselining~\shortcite{wu-iclr18a,spooner-neurips21a}, offline methods through direct value conditioning~\shortcite{tang-darl22a}, and value-based planning through matrix estimation~\shortcite{yang-iclr20b}

Action selection mechanisms can exploit domain knowledge for safety and interpretability~\shortcite{zhang-neurips21ws} or for directed exploration to improve sample efficiency~\shortcite{amin-icml21a}.
Hierarchical settings benefit from latent state decompositions incorporated via modification of the termination condition~\shortcite{harutyunyan-aistats19a}.
Additionally, state-action equivalences help scale Q-learning to large spaces through factorization~\shortcite{lyu-entropy23a}.

\paragraph{Concurrent Optimization.}
Parallelizing optimization using structural decompositions helps with sample efficiency.
Factored MDPs are an excellent way to model factors influencing the content presented to users and help ensemble methods in a parallel regime~\shortcite{tavakol-rs14a}.
Similarly, factored rewards in hierarchical settings help decompose Multi-task problems into a linear combination of individual task MDPs~\shortcite{saxe-icml17a}. 
Alternatively, discretizing continuous sub-actions in multi-dimensional action spaces helps extend the MDP for each sub-action to an undiscounted lower-level MDP, modifying the backup for the Q values using decompositions~\shortcite{metz-arxiv17a}.
Relational decompositions additionally help with masking strategies for Factored Neural Networks~\shortcite{balaji-neurips20a}.

\subsection{Auxiliary Model Pattern} 
\label{sec:patterns:aux-model}

This pattern represents using the structural information in a model.
Using the term model, we specifically refer to methods that utilize a model of the environment to generate experiences, either fully or partially. 
This notion allows us to capture a range of methods, from ones using full-scale world models to generate rewards and next-state transitions to ones that use these methods to generate complete experience sequences.
In our categorization, we specifically look at how the structure is incorporated into such models to help generate some parts of learning experiences.

% \begin{wrapfigure}{r}{0.35\textwidth}
%     \centering
%     \vspace{-10pt}
%     \framebox[0.35\textwidth]{\includegraphics[width=0.2\textwidth, trim={0.6cm 0.1cm 0.4cm 0cm}]{figures/patterns/Aux-Model.pdf}}
%     \vspace{-20pt}
% \end{wrapfigure} 

\begin{figure}[htb]
    \centering
    % \vspace{-10pt}
    \framebox[0.3\textwidth]{\includegraphics[width=0.2\textwidth, trim={0.5cm 0.1cm 0.4cm 0cm}]{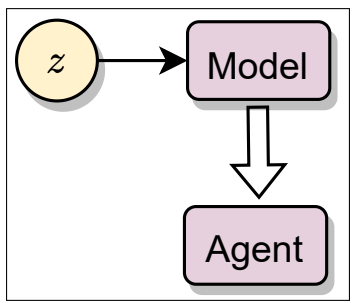}}
    % \vspace{-20pt}
    \caption{Auxiliary Model Pattern}
\end{figure} 

% Example
Our taxi agent could learn a latent model of city traffic based on past experiences. 
This model could be used to plan routes that avoid traffic and reach destinations faster.
Alternatively, the agent could learn an ensembling technique to combine multiple models, each of which model-specific components of the traffic dynamics.
With models, there is usually a trade-off between model complexity and accuracy, and it is essential to manage this carefully to avoid overfitting and maintain robustness.
To this end, incorporating structure helps make the model-learning phase more efficient while allowing reuse for generalization.
Hence, in \cref{fig:pattern-proclivities}, the auxiliary model pattern shows a solid propensity to utilize structure for sample efficiency.
In the following paragraphs, we explicitly discuss models that utilize decompositions and models used for creating decompositions.

% \begin{table}[htbp]
% \caption{Overview of Literature using the Auxiliary Model pattern}    
\begin{center}
    \scriptsize
    \centering
    \begin{tabularx}{\textwidth}{Y|Y|X|X|X|X}
    % \caption{Overview of Literature using the Auxiliary Model pattern}
    % 
    \toprule[1pt]\midrule[0.3pt]
    \textbf{Space}              &  \textbf{Type}            & \multicolumn{1}{c}{\textbf{Efficiency}}  & \multicolumn{1}{c}{\textbf{Generalization}} & \multicolumn{1}{c}{\textbf{Interpretabiltiy}} & \multicolumn{1}{c}{\textbf{Safety}} \\  
    \hline
    \multirow{3}{*}{Goals}      & Factored                  & & \shortciteA{ding-neurips22a} & & \\
                                \cline{2-6}
                                & Relational               & & \shortciteA{sohn-neurips18a}, \shortciteA{sohn-iclr20a} & & \\
                                \cline{2-6}
                                & Modular                   & \shortciteA{icarte-jair22a} & \shortciteA{icarte-jair22a} & \shortciteA{icarte-jair22a} & \\
    \hline
    % \midrule[0.3pt]
    \multirow{4}{*}{States}     & Latent                    & \shortciteA{gasse-arxiv21a}, \shortciteA{wang-icml22b}, \shortciteA{hafner-arxiv23a}, \shortciteA{vanderpol-aamas20a}, \shortciteA{ghorbani-arxiv20a}, \shortciteA{tsividis-arxiv21a}, \shortciteA{yin-arxiv23a} & \shortciteA{vanderpol-aamas20a}, \shortciteA{wang-icml22b}, \shortciteA{hafner-arxiv23a}, \shortciteA{hafner-icml20a}, \shortciteA{zhang-iclr21b}, \shortciteA{tsividis-arxiv21a}& & \shortciteA{simao-aamas21a} \\
                                \cline{2-6}
                                & Factored                  & \shortciteA{innes-uai20a}, \shortciteA{seitzer-neurips21a}, \shortciteA{andersen-nips17a}, \shortciteA{ross-uai08}, \shortciteA{singh-icml21a}, \shortciteA{pitis-neurips20a} & \shortciteA{young-icml23a}, \shortciteA{ding-neurips22a} & & \\
                                \cline{2-6}
                                & Relational                & \shortciteA{chen-iros20a}, \shortciteA{biza-iclr22a},  \shortciteA{biza-arxiv22b}, \shortciteA{kipf-iclr20a}, \shortciteA{tsividis-arxiv21a}, \shortciteA{singh-icml21a}, \shortciteA{pitis-neurips20a} & \shortciteA{biza-iclr22a}, \shortciteA{biza-arxiv22b}, \shortciteA{veerapaneni-corl20a}, \shortciteA{kipf-iclr20a}, \shortciteA{tsividis-arxiv21a} & \shortciteA{xu-axiv21a} & \\
                                \cline{2-6}
                                & Modular                   & \shortciteA{abdulhai-aaai22a}, \shortciteA{andersen-nips17a}, \shortciteA{icarte-jair22a}, \shortciteA{blanco-jair21a} & \shortciteA{icarte-jair22a}, \shortciteA{blanco-jair21a} & \shortciteA{icarte-jair22a}, \shortciteA{blanco-jair21a} & \\
    \hline
    % \midrule[0.3pt]
    \multirow{3}{*}{Actions}    & Latent                    & \shortciteA{vanderpol-aamas20a} &\shortciteA{vanderpol-aamas20a} & & \\ 
                                \cline{2-6}
                                & Factored                  & \shortciteA{spooner-neurips21a}, \shortciteA{geiser-socs20a}, \shortciteA{innes-uai20a}, \shortciteA{pitis-neurips20a} & \shortciteA{ding-neurips22a} & & \\ 
                                \cline{2-6}
                                & Relational                & \shortciteA{biza-iclr22a}, \shortciteA{pitis-neurips20a} & \shortciteA{biza-iclr22a}  & & \\ 
                                \cline{2-6}
                                & Modular                   & \shortciteA{blanco-jair21a}, \shortciteA{yang-ijcai18a} & \shortciteA{blanco-jair21a} & \shortciteA{blanco-jair21a} & \\
    \hline
    % \midrule[0.3pt]
    \multirow{2}{*}{Rewards}    & Latent                    & \shortciteA{vanderpol-aamas20a} & \shortciteA{zhang-iclr21b}, \shortciteA{vanderpol-aamas20a}, \shortciteA{lee-neurips21a}, \shortciteA{sohn-neurips18a}, \shortciteA{sohn-iclr20a} & & \\ 
                                \cline{2-6}
                                & Factored                  & & \shortciteA{sohn-neurips18a} & & \shortciteA{wang-aaai20a}, \shortciteA{baheri-arxiv20a} \\ 
    \hline
    % \midrule[0.3pt]
    \multirow{2}{*}{Dynamics}  & Latent                    & \shortciteA{woo-aaai22a}, \shortciteA{fu-icml21a}, \shortciteA{vanderpol-aamas20a}, \shortciteA{wang-icml22a} & \shortciteA{zhang-iclr21b}, \shortciteA{woo-aaai22a}, \shortciteA{vanderpol-aamas20a}, \shortciteA{fu-icml21a}, \shortciteA{guo-iclr22a}, \shortciteA{wang-icml22a} &  & \\ 
                            \cline{2-6}
                            & Factored                  & \shortciteA{fu-icml21a}, \shortciteA{schiewer-lod21a} & \shortciteA{goyal-iclr21a}, \shortciteA{fu-icml21a} & \shortciteA{schiewer-lod21a}, \shortciteA{kaiser-esann19a} & \\ 
                            \cline{2-6}
                            & Relational                & \shortciteA{buesing-iclr19a} & & \shortciteA{rossum-arxiv21a} & \\
                            \cline{2-6}
                            & Modular                   & \shortciteA{abdulhai-aaai22a}, \shortciteA{wu-aamas19a}, \shortciteA{wen-neurips20a} & \shortciteA{wu-aamas19a} & & \\ 
    \midrule[0.3pt]\bottomrule[1pt]
    \caption{Overview of Literature using the Auxiliary Model pattern}
    \end{tabularx}
    \label{table:patterns:aux-model}
\end{center}
% \end{table}

\paragraph{Models with structured representations.}
\shortciteA{young-icml23a} utilize factored decomposition for state space to demonstrate the benefits of model-based methods in combinatorially complex environments.
Similarly, the dreamer models~\shortcite{hafner-icml20a,hafner-arxiv23a} utilize latent representations of pixel-based environments.

Object-oriented representations for states help bypass the need to learn latent factors using CNNs in MBRL~\shortcite{biza-arxiv22b} or as random variables whose posterior can be refined using NNs~\shortcite{veerapaneni-corl20a}.
Graph (Convolutional) Networks~\shortcite{zhang-csn19a} capture rich higher-order interaction data, such as crowd navigation~\shortcite{chen-iros20a}, or invariances~\shortcite{kipf-iclr20a}.
Action equivalences help learn latent models (Abstract MDPs)~\shortcite{vanderpol-aamas20a} for planning and value iteration.

\paragraph{Models for task-specific decompositions.}
Another way to utilize decompositions in models is to capture task-specific decompositions.
Models that capture some form of irrelevance, such as observational and interventional data in Causal RL~\shortcite{gasse-arxiv21a}, or task-relevant vs. irrelevant data~\shortcite{fu-icml21a} help with generalization and sample efficiency gains.
Latent representations help models capture control-relevant information~\shortcite{wang-icml22b} or subtask dependencies~\shortcite{sohn-neurips18a}.

Models for safety usually incorporate some measure of cost to abstract safe states~\shortcite{simao-aamas21a}, or unawareness to factor states and actions~\shortcite{innes-uai20a}.
Alternatively, models can also directly guide exploration mechanisms through latent causal decompositions~\shortcite{seitzer-neurips21a} and state subspaces~\cite{ghorbani-arxiv20a} to gain sample efficiency.
Generative methods such as CycleGAN~\shortcite{zhu-ieeecv17a} are also excellent ways to use Latent models of different components of an MDP to generate counterfactual trajectories~\shortcite{woo-aaai22a}.

\subsection{Portfolio Pattern} 
\label{sec:patterns:portfolio}

This pattern uses structural information to create a database, or a warehouse, of entities that can be combined to achieve a specific objective. 
These can be learned policies and value functions or even models. 
Given the online nature of such methods, they are often targeted toward continual and life-long learning problems. 
The inherent modularity in such methods often leads them to focus on knowledge reuse as a central theme. 

% \begin{wrapfigure}{r}{0.35\textwidth}
%     \centering
%     \vspace{-10pt}
%     \framebox[0.35\textwidth]{\includegraphics[width=0.35\textwidth, trim={0.6cm 0cm 0.2cm 0cm}]{figures/patterns/Subspace.pdf}}
% \end{wrapfigure}

\begin{figure}[htb]
    \centering
    % \vspace{-10pt}
    \framebox[0.35\textwidth]{\includegraphics[width=0.35\textwidth, trim={0.6cm 0cm 0.2cm 0cm}]{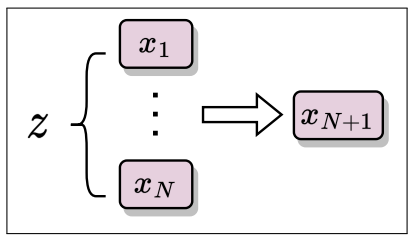}}
    \caption{Portfolio Pattern}
\end{figure}

%Example
The taxi from our running example could maintain a database of value functions or policies for different parts of the city or at different times of the day. These could be reused as the taxi navigates through the city, making learning more efficient.
% Downside
Portfolios generally improve efficiency and generalization and have been traditionally implemented through the skills and options frameworks. 
An essential consideration in this pattern is managing the portfolio's size and diversity to avoid biasing the learning process too much toward past experiences. 

So far, the portfolios have primarily been applied to sample efficiency and generalization.
However, they also overlap with interpretability since the stored data can be easily used to analyze the agent's behavior and understand the policy for novel scenarios.
Consequently, these objectives are equitably distributed in \cref{fig:pattern-proclivities}.  

\begin{center}
    \scriptsize
    \centering
    \begin{tabularx}{\textwidth}{Y|Y|X|X|X|X}
    \toprule[1pt]\midrule[0.3pt]    
    \textbf{Space}             & \textbf{Type}             & \multicolumn{1}{c}{\textbf{Efficiency}}  & \multicolumn{1}{c}{\textbf{Generalization}} & \multicolumn{1}{c}{\textbf{Interpretabiltiy}} & \multicolumn{1}{c}{\textbf{Safety}}\\  
    \hline
    \multirow{3}{*}{Goals}     & Factored                  & & \shortciteA{mendez-iclr22a}, \shortciteA{devin-icra17a} & & \\ 
                               \cline{2-6}            
                               & Relational                & & & \shortciteA{prakash-arxiv22a} &\\
                               \cline{2-6}
                               & Modular                   & \shortciteA{gehring-neurips21a} & \shortciteA{mendez-iclr22a} & \shortciteA{prakash-arxiv22a} & \\
    \hline
    % \midrule[0.3pt]
    \multirow{3}{*}{States}    & Latent                    & & \shortciteA{hu-arxiv19a}, \shortciteA{bhatt-neurips22a} & & \\
                               \cline{2-6}
                               & Factored                  & \shortciteA{mankowitz-arxiv15a}, \shortciteA{yarats-icml21a} & \shortciteA{mendez-iclr22a}, \shortciteA{goyal-iclr20a}, \shortciteA{yarats-icml21a} & & \\ 
                               \cline{2-6}            
                               & Modular                   & \shortciteA{blanco-jair21a} & \shortciteA{mendez-iclr22a}, \shortciteA{goyal-iclr20a}, \shortciteA{blanco-jair21a} & \shortciteA{blanco-jair21a} & \\
    \hline
    % \midrule[0.3pt]
    \multirow{2}{*}{Actions}   & Latent                    & & \shortciteA{gupta-iclr17a} & & \\
                               \cline{2-6}
                               & Modular                   & \shortciteA{li-robio18a}, \shortciteA{blanco-jair21a}, \shortciteA{devin-neurips19a} & \shortciteA{blanco-jair21a}, \shortciteA{devin-neurips19a}, \shortciteA{nam-iclr22a}, \shortciteA{peng-neurips19a}, \shortciteA{barreto-neurips19a}, \shortciteA{sharma-iclr20a}, \shortciteA{xu-neurips20a} & \shortciteA{blanco-jair21a} &  \\
                              
    \hline
    Rewards                    & Factored                  & & \shortciteA{haarnoja-icra18a}, \shortciteA{mendez-iclr22a}, \shortciteA{gaya-darl22a}, \shortciteA{gaya-iclr22a}  & & \\
    \hline
    % \midrule[0.3pt]
    \multirow{2}{*}{Dynamics}  & Latent                    & & \shortciteA{bhatt-neurips22a} & &\\
                               \cline{2-6}
                               & Factored                  & \shortciteA{shyam-icml19a}, \shortciteA{schiewer-lod21a} & \shortciteA{devin-icra17a}, \shortciteA{mendez-iclr22a}  & \shortciteA{schiewer-lod21a} & \\ 
                               \cline{2-6}            
                               & Modular                   & \shortciteA{wu-aamas19a} & \shortciteA{gaya-darl22a}, \shortciteA{gaya-iclr22a}, \shortciteA{mendez-iclr22a}, \shortciteA{wu-aamas19a}  & & \\
    \hline
    % \midrule[0.3pt]
    \multirow{2}{*}{Policies}  & Latent                    & & \shortciteA{gupta-iclr17a} & \shortciteA{verma-icml18}& \\
                               \cline{2-6}
                               & Modular                   & \shortciteA{wolf-axiv23a}, \shortciteA{florensa-iclr17a}, \shortciteA{heess-arxiv16a}, \shortciteA{eysenbach-iclr19a}, \shortciteA{raza-icit19a}, \shortciteA{mankowitz-arxiv15a}, \shortciteA{mendez-neurips20a}, \shortciteA{hausman-iclr18a} & \shortciteA{florensa-iclr17a}, \shortciteA{heess-arxiv16a}, \shortciteA{mendez-neurips20a}, \shortciteA{kaplanis-icml19a}, \shortciteA{hausman-iclr18a} & \shortciteA{verma-icml18}& \\ 
    \midrule[0.3pt]\bottomrule[1pt]
    \caption{Survey of literature utilizing the portfolio pattern}
    \end{tabularx}
    \label{table:patterns:portfolio}
    \vspace{-20pt}
\end{center}

\paragraph{Policy portfolios.}
Policy subspaces~\shortcite{gaya-iclr22a} utilize shared latent parameters in policies to learn a subspace, the linear combinations of which help create new policies. 
Extending these subspaces by warehousing additional policies naturally extends them to continual settings~\shortcite{gaya-darl22a}.

Using goals and rewards, task factorization endows warehousing policies and Q values in multi-task lifelong settings. 
Relationship graphs between existing tasks generated from a latent space provide a way to model lifelong multi-task learning problems~\shortcite{mendez-iclr22a}.
On the other hand, methods such as those presented by \shortciteA{devin-icra17a} factor MDPs into agent-specific and task-specific degrees of variation, for which individual modules are trained.
Disentanglement using variational encoder-decoder models~\shortcite{hu-arxiv19a} helps control morphologically different agents by factorizing dynamics into shared and agent-specific factors.
Additionally, methods similar to the work of \shortciteA{raza-icit19a} partition the agent's problem into interconnected sub-agents that learn local control policies.

Methods that utilize the skills framework effectively create a portfolio of learned primitives, similar to options in HRL. 
These are subsequently used for maximizing mutual information in lower layers~\shortcite{florensa-iclr17a}, sketching together a policy~\shortcite{heess-arxiv16a}, diversity-seeking priors in continual settings~\shortcite{eysenbach-iclr19a}, or for partitioned states spaces~\shortcite{mankowitz-arxiv15a}.
Similarly, \shortciteA{gupta-iclr17a} apply the portfolio pattern on a latent embedding space, learned using auxiliary optimization.

% \shortciteA{gehring-neurips21a}, in addition to auxiliary optimization and augmentation, also portfolio their learned policies for efficient exploration.

\paragraph{Decomposed Models.}
Decompositions that inherently exist in models lead to approaches that often ensemble multiple models that individually reflect different aspects of the problem. 
Ensemble methods such as Recurrent Independent Mechanisms~\shortcite{goyal-iclr21a} capture the dynamics in individual modules that sparsely interact and use attention mechanisms~\shortcite{vaswani-neurips17a}. 
Ensembling dynamics also helps with few-shot adaptation to unseen MDPs~\shortcite{lee-neurips21a}.
Factored models combined with relational decompositions help bind actions to object-centric representations~\shortcite{biza-iclr22a}.
Latent representations in hierarchical settings~\shortcite{abdulhai-aaai22a} improve the sample inefficiency of methods such as the Deep Option Critic~\shortcite{bacon-aaai17a}.

\subsection{Environment Generation Pattern}
\label{sec:patterns:env-gen}

This pattern uses structural information to create task, goal, or dynamics distributions from which MDPs can be sampled. 
This subsumes the idea of procedurally generated environments while incorporating methods that use auxiliary models to induce structure in the environment generation process.
The decomposition is reflected in the aspects of the environment generation that are impacted by the generative process, such as dynamics, reward structure, state space, etc.
Given the online nature of environment generation, methods in this pattern address curriculum learning in one way or another.

% \begin{wrapfigure}{r}{0.35\textwidth}
%     \centering
%     \vspace{-10pt}
%     \framebox[0.35\textwidth]{\includegraphics[width=0.3\textwidth, trim={0.3cm 0.05cm 0.3cm 0cm}]{figures/patterns/Env-Gen.pdf}}
%     \vspace{-20pt}
% \end{wrapfigure} 

\begin{figure}[htb]
    \centering
    % \vspace{-10pt}
    \framebox[0.35\textwidth]{\includegraphics[width=0.3\textwidth, trim={0.3cm 0.05cm 0.3cm 0cm}]{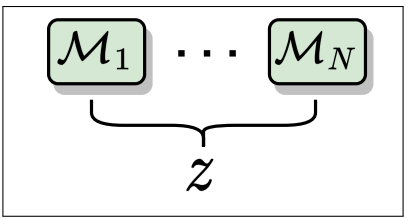}}
    % \vspace{-20pt}
    \caption{Environment Generation Pattern}
\end{figure} 

% Examples
In the taxi example, a curriculum of tasks could be generated, starting with simple tasks (like navigating an empty grid) and gradually introducing complexity (like adding traffic and passengers with different destinations).
Ensuring that the generated MDPs provide good coverage of the problem space is crucial to avoid overfitting to a specific subset of tasks.
This necessitates additional diversity constraints that must be incorporated into the environment generation process.
Structure, crucially, provides additional interpretability and controllability in the environment generation process, thus making benchmarking easier than methods that use unsupervised techniques~\shortcite{laskin-neurips21a}.

\begin{center}
    \scriptsize
    \centering
    % \caption{Overview of Literature using the Warehouse pattern}
     
    \begin{tabularx}{\textwidth}{Y|Y|X|X|X|X}
    \toprule[1pt]\midrule[0.3pt]    
    \textbf{Space}             & \textbf{Type}             & \multicolumn{1}{c}{\textbf{Efficiency}}  & \multicolumn{1}{c}{\textbf{Generalization}} & \multicolumn{1}{c}{\textbf{Interpretabiltiy}} & \multicolumn{1}{c}{\textbf{Safety}}\\  
    \hline
    \multirow{2}{*}{Goals}     & Relational                & \shortciteA{illanes-icaps20a}, \shortciteA{gur-neurips21a} & \shortciteA{kumar-neurips22a} & \shortciteA{gur-neurips21a} & \\ 
                               \cline{2-6}            
                               & Modular                   & \shortciteA{Kulkarni-neurips16}, \shortciteA{illanes-icaps20a} & \shortciteA{narvekar-aamas16a},  \shortciteA{mendez-collas22} & & \\
                               
    \hline
    % \midrule[0.3pt]
    \multirow{3}{*}{States}    & Latent                    & & \shortciteA{wang-neurips21a}, \shortciteA{bhatt-neurips22a} & & \\ 
                               \cline{2-6}
                               & Factored                  & \shortciteA{lu-arxiv18a}, \shortciteA{mirsky-darl22a} & \shortciteA{mirsky-darl22a} &\shortciteA{lu-arxiv18a}, \shortciteA{mirsky-darl22a} & \\ 
                               \cline{2-6}            
                               & Relational                & \shortciteA{lu-arxiv18a}, \shortciteA{ada-arxiv23a} & \shortciteA{ada-arxiv23a} & \shortciteA{lu-arxiv18a}& \\
                            
    \hline
    % \midrule[0.3pt]
    \multirow{4}{*}{Rewards}   & Latent                    & & \shortciteA{wang-neurips21a}, \shortciteA{lee-neurips21a} & & \\
                               \cline{2-6}            
                               & Factored                  & \shortciteA{chu-arxiv23a} & \shortciteA{mendez-collas22} & & \\
    \hline
    % \midrule[0.3pt]
    \multirow{4}{*}{Dynamics}  & Latent                    & & \shortciteA{kumar-iclr21a}, \shortciteA{bhatt-neurips22a} & & \\
                               \cline{2-6}            
                               & Factored                  & \shortciteA{chu-arxiv23a}, \shortciteA{mirsky-darl22a} & \shortciteA{mirsky-darl22a}, \shortciteA{narvekar-aamas16a}, \shortciteA{mendez-collas22} & \shortciteA{mirsky-darl22a} & \\
                               \cline{2-6}
                               & Relational                & \shortciteA{illanes-icaps20a},  \shortciteA{ada-arxiv23a} & \shortciteA{wang-neurips21a},  \shortciteA{ada-arxiv23a} & \shortciteA{wang-neurips21a},  \shortciteA{ada-arxiv23a} & \\
                               \cline{2-6}
                               & Modular                   & \shortciteA{illanes-icaps20a}, \shortciteA{mirsky-darl22a} & \shortciteA{mirsky-darl22a} & \shortciteA{mirsky-darl22a} & \\
    \midrule[0.3pt]\bottomrule[1pt]
    \caption{Survey of literature utilizing the environment generation pattern}
    \end{tabularx}
    \label{table:patterns:env-gen}
\end{center}

Rule-based grammars help model the compositional nature of learning problems.
\shortciteA{kumar-iclr21a} utilize this to impact the transition dynamics and generate environments.
This allows them to train agents with an implicit compositional curriculum. 
\shortciteA{kumar-neurips22a} use these grammars in their auxiliary optimization procedure.
Another way to capture task dependencies is through latent graphical models, which generate the state-space, reward functions, and transition dynamics~\shortcite{wang-neurips21a,ada-arxiv23a}.

Latent dynamics models allow simulating task distributions, which help with generalization~\shortcite{lee-neurips21a}.
Clustering methods such as Exploratory Task Clustering~\shortcite{chu-arxiv23a} explore task similarities by meta-learning a clustering method through an exploration policy. 
In some sense, they recover a factored decomposition on the task space where individual clusters can be further used for policy adaptation.

\subsection{Explicitly Designed}
\label{sec:patterns:explicit-design}

This pattern encompasses all methods where the inductive biases manifest in specific architectures or setups that reflect the decomposability of the problem that they aim to utilize. 
Naturally, this includes specifically designed neural architectures and extends to other methods, such as sequential architectures, to capture hierarchies, relations, etc. 
Crucially, the usage of structural information is limited to the specificity of the architecture and not any other part of the pipeline.

% \begin{wrapfigure}{r}{0.35\textwidth}
%     \centering
%     \vspace{-10pt}
%     \framebox[0.35\textwidth]{\includegraphics[width=0.3\textwidth, trim={0.8cm 0.5cm 0.8cm 0.3cm}]{figures/patterns/Exp-Design.pdf}}
%     \vspace{-20pt}
% \end{wrapfigure}

\begin{figure}[htb]
    \centering
    % \vspace{-10pt}
    \framebox[0.35\textwidth]{\includegraphics[width=0.3\textwidth, trim={0.8cm 0.5cm 0.8cm 0.3cm}]{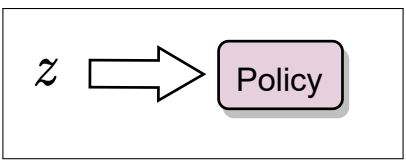}}
    % \vspace{-20pt}
    \caption{Explicitly Designed Pattern}
\end{figure}

% Example
In the case of the taxi, a neural architecture could be designed to process the city grid as an image and output a policy. 
Techniques like convolutional layers could be used to capture the spatial structure of the city grid. 
Different network parts could be specialized for different subtasks, like identifying passenger locations and planning routes.
%&downside
However, this pattern involves a considerable amount of manual tuning and experimentation, and it's critical to ensure that these designs generalize well across different tasks. 
Designing specific neural architectures can provide better interpretability, enabling the ability to decompose different components and simulate them independently. 
Consequently, this pattern shows the highest proclivity to interpretability, with Generalization being a close second in \cref{fig:pattern-proclivities}. 

\begin{center}
    \scriptsize
    \centering
    % \caption{Overview of Literature using the Warehouse pattern}
     
    \begin{tabularx}{\textwidth}{Y|Y|X|X|X|X}
    \toprule[1pt]\midrule[0.3pt]    
    \textbf{Space}             & \textbf{Type}             & \multicolumn{1}{c}{\textbf{Efficiency}}  & \multicolumn{1}{c}{\textbf{Generalization}} & \multicolumn{1}{c}{\textbf{Interpretabiltiy}} & \multicolumn{1}{c}{\textbf{Safety}}\\  
    \hline
    \multirow{2}{*}{Goals}     & Factored                  & \shortciteA{zhou-arxiv22a} & \shortciteA{zhou-arxiv22a} & \shortciteA{alabdulkarim-arxiv22a} & \\ 
                               \cline{2-6}            
                               & Relational                & \shortciteA{zhou-arxiv22a} & \shortciteA{zhou-arxiv22a} & & \\
                               
    \hline
    % \midrule[0.3pt]
    \multirow{4}{*}{States}    & Latent                    & \shortciteA{wang-icarcv16a} & \shortciteA{yang-neurips20a} & & \\
                               \cline{2-6}
                               & Factored                  & \shortciteA{zhou-arxiv22a} & \shortciteA{zhou-arxiv22a} & & \\
                               \cline{2-6}
                               & Relational                & \shortciteA{zhou-arxiv22a},\shortciteA{mambelli-iclr22a},\shortciteA{shanahan-icml20a},\shortciteA{zambaldi-iclr19a} & \shortciteA{zhou-arxiv22a},\shortciteA{mambelli-iclr22a},\shortciteA{shanahan-icml20a},\shortciteA{zambaldi-iclr19a},\shortciteA{lampinen-icml22a}, \shortciteA{sharma-uai22a} & \shortciteA{zambaldi-iclr19a}, \shortciteA{payani-arxiv20a}  & \\
                               \cline{2-6}
                               & Modular                   & & & & \\
    \hline
    % \midrule[0.3pt]
    \multirow{3}{*}{Actions}   & Latent                    & \shortciteA{wang-icarcv16a} & & & \\
                               \cline{2-6}
                               & Factored                  & \shortciteA{tavakoli-aaai18a} & & \shortciteA{tavakoli-aaai18a} & \\
                               \cline{2-6}
                               & Relational                & \shortciteA{garg-icml20a} & & \shortciteA{garg-icml20a} & \\
                              
    \hline
    % \midrule[0.3pt]
    \multirow{1}{*}{Rewards}   & Latent                    & & \shortciteA{yang-neurips20a} & & \\
                                \cline{2-6}            
                               & Factored                   & & & & \shortciteA{baheri-arxiv20a}\\
    \hline
    % \midrule[0.3pt]
    \multirow{2}{*}{Dynamics}  & Latent                    & \shortciteA{vanderpol-neurips20a} & \shortciteA{deramo-iclr20a}, \shortciteA{guo-iclr22a} & & \\
                               \cline{2-6}
                               & Factored                  & \shortciteA{srouji-icml18a},\shortciteA{hong-iclr22a} & & & \\
                               \cline{2-6}
                                & Relational               & & \shortciteA{lampinen-icml22a} & & \\ 
    \hline
    % \midrule[0.3pt]
    Policies                   & Relational                & \shortciteA{olivia-ijcnn22a}, \shortciteA{garg-icml20a} & \shortciteA{wang-iclr18a} & \shortciteA{garg-icml20a} & \\
                               \cline{2-6}
                               & Modular                   & & \shortciteA{shu-iclr18a} &  \shortciteA{shu-iclr18a},\shortciteA{mu-tmlr22a} & \\ 
    \midrule[0.3pt]\bottomrule[1pt]
    \caption{Survey of literature utilizing the explicitly designed pattern}
    \end{tabularx}
    \label{table:patterns:exp-designed}
    \vspace{-20pt}
\end{center}

\paragraph{Splitting Functionality.}
One way to bias the architecture is to split its functionality into different parts.
Most of the works that achieve such disambiguation are either Factored or Relational.
Structured Control Nets~\shortcite{srouji-icml18a} model linear and non-linear aspects of the dynamics individually and combine them additively to gain sample efficiency and generalization.
Alternatively, Bi-linear Value Networks~\shortcite{hong-iclr22a} architecturally decompose dynamics into state and goal-conditioned components to produce a goal-conditioned Q-function.
Action Branching architectures~\shortcite{tavakoli-aaai18a} used a shared representation that is then factored into separate action branches for individual functionality.
This approach bears similarity to capturing multi-task representations using bottlenecks~\shortcite{deramo-iclr20a}.

Relational and Modular biases manifest in hierarchical architectures. 
This also allows them to add more interpretability to the architecture. 
Two-step hybrid policies~\shortcite{mu-tmlr22a}, for example, demonstrate an explicit method to make policies more interpretable through splitting actions into pruners and selector modules. 
On the other hand, routing hierarchies explicitly capture modularity using sub-modules that a separate policy can use for routing them~\shortcite{shu-iclr18a,yang-neurips20a}.

\paragraph{Capturing Invariances in Architectures.}
Specialized architectures also help capture invariance in the problem. 
Symbolic Networks~\shortcite{garg-icml20a,sharma-uai22a} train a set of shared parameters for Relational MDPs by first converting them to Graphs and then capturing node embeddings using Neural Networks.
Homomorphic Networks~\shortcite{vanderpol-neurips20a} capture symmetry into specialized MLP and CNN architectures.
An alternate approach to incorporating symmetry is through basis functions~\shortcite{wang-icarcv16a}.

Attention mechanisms explicitly help capture entity-factored scenarios~\shortcite{janisch-arxiv20a,shanahan-icml20a,zhou-arxiv22a}. 
Relational and Graph Networks capture additional relational inductive biases explicitly. 
Linear Relation Networks~\shortcite{mambelli-iclr22a} provides an architecture that scales linearly with the number of objects.
Graph networks have also been used to model an agent's morphology in embodied control explicitly~\shortcite{wang-iclr18a,olivia-ijcnn22a}.

\paragraph{Specialized Modules.}
These are a class of methods that combines the best of both worlds by capturing invariance in additional specialized modules.
Such modules capture relational structure in semantic meaning~\shortcite{lampinen-icml22a}, relational encoders for auxiliary models~\shortcite{guo-iclr22a}, or specialized architectures for incorporating domain knowledge~\shortcite{payani-arxiv20a}.
\section{Open Problems in Structured Reinforcement Learning}
\label{sec:connection}

Having discussed our patterns-oriented framework for understanding how to incorporate structure into the RL pipeline, we now turn to connect our framework with existing sub-fields of RL. 
We examine existing paradigms in these sub-fields from two major perspectives: \emph{Scalability and Robustness}. 
Sparse data scenarios require more intelligent ways to use limited experiences. 
In contrast, abundant data scenarios might suffer from data quality since they might be generated from noisy and often unreliable sources. 
% These dimensions serve as a canvas upon which we can position and understand different RL paradigms in sub-fields such as Offline RL,  Unsupervised RL, Foundation Models in RL, Partial observability, Big Worlds, Automated RL, and Meta-RL. 

% \paragraph{Data Dependency} measures how much a method depends on the volume and heterogeneity of data in the form of reward and collected trajectories. 
% Under this view, the quality of the available data depends on how well it allows a pipeline to learn an optimal policy.
% The quality can be affected by the hardness of exploration, in which case, the sparsity and deceptiveness of the reward prohibit an agent from visiting states beneficial to learning the optimal policy. 
% Alternatively, even in environments where the reward is not sparse, the available data can be generated from noisy and unreliable sources, making it hard to learn the optimal policy.
% Thus,

\paragraph{Scalability} measures how methods scale with the increasing problem complexity in terms of the size of the state and action spaces, complex dynamics, noisy reward signals, and longer task horizons. 
On the one hand, methods might specifically require low dimensional spaces and might not scale so well with increasing the size of these spaces, and on the other, some methods might be overkill for simple problems but better suited for large spaces.

\paragraph{Robustness} measures the response of methods to changes in the environment. 
While the notion overlaps with generalization, robustness for our purposes more holistically looks at central properties of data distribution, such as initial state distributions and multi-modal evaluation returns. 
Under this notion, fundamentally different learning dynamics might be robust to different kinds of changes in the environment.
% While more robust methods could, in principle, be more generalizable, other factors could limit such generalization.

\paragraph{Structure of the Section.} 
In the following subsections, we cover sub-fields of RL that lie at different areas of the Scalability and Robustness space. 
Each subsection covers an existing sub-field and its challenges.
We then present some examples in which our framework can bolster further research and practice in these fields. 
Finally, we collate this discussion into takeaways that can be combined into specific areas of further research.
These are shown in the blue boxes at the end of each section.

\subsection{Offline RL}
Offline Reinforcement Learning (also known as batch RL)~\shortcite{prudencio-itnnls23a} involves learning from a fixed dataset without further interaction with the environment. 
This approach can benefit when active exploration is costly, risky, or infeasible. 
Consequently, such methods are highly data-dependent due to their reliance on the collected dataset, and they do not generalize well due to the limitations of the pre-collected data. 
The three dominant paradigms in Offline RL -- Behavior Cloning~\shortcite{bain-mi95a}, Q-Learning~\shortcite{kumar-neurips20a}, and Sequence Modelling~\shortcite{chen-neurips21b} -- uniformly degrade in performance as the state-space increases~\shortcite{bhargava-arxv23a}.
Offline RL has challenges, including effectively overcoming distributional shifts and exploiting the available dataset.
Structural decomposition can be crucial in addressing these challenges in different ways, as summarized in the following paragraphs.

\paragraph{Improved Exploitation of Dataset.} 
Task decomposition allows learning individual policies or value functions for different subtasks, which could potentially leverage the available data more effectively. 
% Task decompositions, thus, open up new avenues for research in sample-efficient learning algorithms. 
For example, modular decomposition utilizing portfolios of policies for individual modules using the corresponding subset of the data might be more sample-efficient than learning a single policy for the entire task. 
Task decompositions, thus, open up new avenues for developing specialized algorithms that effectively learn from limited data about each subtask while balancing the effects of learning different subtasks. 
Practitioners can leverage such decompositions to maximize the utility of their available datasets by training models that effectively handle specific subtasks, potentially improving the overall system's performance with the same dataset.

\paragraph{Mitigating Distributional Shift.} 
The structural information could potentially help mitigate the effect of distributional shifts. 
For instance, if some factors are less prone to distributional shifts in a factored decomposition, we could focus more on those factors during learning. This opens up venues for gaining theoretical insights into the complex interplay of structural decompositions, task distributions, and policy performance. 
On the other hand, practical methods for environments where distributional shifts are standard could leverage structural decomposition to create more robust RL systems.

\paragraph{Auxiliary Tasks for Exploration.} 
Structural decomposition can be used to define auxiliary tasks that facilitate learning from the dataset. 
For instance, in a relational decomposition, we could define auxiliary tasks that involve predicting the relationships between different entities, which could help in learning a valuable representation of the data.
Using the proposed framework, researchers can explore how to define meaningful auxiliary tasks that help the agent learn a better representation of the environment.
This could lead to new methods that efficiently exploit the available data by learning about these auxiliary tasks. 
Practitioners can design auxiliary tasks based on the specific decompositions of their problem. 
For example, if the task has a clear relational structure, auxiliary tasks that predict the relations between different entities can potentially improve the agent's understanding of the environment and its overall performance.

\begin{mybox}
    \textbf{Research Area 1: Structured Offline RL}
    \begin{itemize}
        \item Use a Factored or Relational decomposition to create abstractions that can help with distribution shift and auxiliary interpretability.
        \item Implement a Modular design with each module targeting a specific sub-problem, improving Scalability. 
        \item Employ policy reuse by policy portfolios learned for sub-problems across tasks.
        \item If sufficient interaction data is available, employ data augmentation strategies for counterfactual scenarios using latent models.
    \end{itemize}
\end{mybox}

\subsection{Unsupervised RL}
Unsupervised RL~\shortcite{laskin-neurips21a} refers to the sub-field of behavior learning in RL, where an agent learns to interact with an environment without receiving explicit feedback or guidance in the form of rewards. 
Methods in this area can be characterized based on the nature of the metrics that are used to evaluate performance intrinsically~\shortcite{srinivas-misc21a}.
\emph{Knowledge-based} methods define a self-supervised task by making predictions on some aspect of the environment~\shortcite{pathak-icml17a,chen-arxiv22a,chen-jcc22a}, \emph{Data-based methods} maximize the state visitation entropy for exploring the environment~\shortcite{hazan-icml19a,zhang-aaai21a,guo-arxiv21a,mutti-aaai21a,mutti-aaai22a}, and \emph{Competence-based} methods maximize the mutual information between the trajectories and space of learned skills~\shortcite{mohamed-neurips15a,gregor-arxiv16a,baumli-aaai21a,jiang-neurips22a,zeng-ieee22a}.
The pre-training phase allows these methods to learn the underlying structure of data.
% and, thus, allows these methods to work well with differing data qualities.
However, this phase also requires large amounts of data and, thus, impacts the scalability of such methods for problems where the learned representations are not very useful.
Consequently, such methods currently handle medium complexity problems, with the avenue of better scalability being a topic of further research.

Structural decompositions can help such methods by improving the pre-training phase's tractability and the fine-tuning phase's generality.
Latent decompositions could help exploit structure in unlabeled data, while relational decompositions could add interpretability to the learned representations. 
Through augmentation, conditioning policies on specific parts of the state space can reduce the data needed for fine-tuning.
Additionally, understanding problem decomposition can simplify complex problems into more manageable sub-problems, effectively reducing the perceived problem complexity while incorporating such decomposition in external curricula for fine-tuning.
Incorporating portfolios guided by decompositions for competence-based methods can boost the fine-tuning process of the learned skills.

\begin{mybox}
    \textbf{Research Area 2: Structured Unsupervised RL}
    \begin{itemize}
        \item Use latent decompositions to extract structure from unlabeled data, reducing Data Dependency.
        \item Employ factored and modular decompositions and abstractions to manage scalability by focusing learning on different parts of the problem independently.
        \item Store skills in a portfolio across different modular sub-problems to reuse solutions and enhance Generality.
        \item Manage Problem Complexity by leveraging problem decomposition to simplify the learning task and using decompositions for fine-tuning using curriculum learning.
    \end{itemize}
\end{mybox}

\subsection{Big Data and Foundation models in RL}
Foundation models~\shortcite{brown-neurips20a,openai-arxiv23a,kirillov-arxiv23a} refer to a paradigm where a large model is pre-trained on large and heterogeneous datasets and fine-tuned for specific tasks. 
These models are ``foundational" in the sense that they can serve as the basis for a wide range of tasks, reducing the need for training separate models for each task from scratch. 

Foundation models for RL come increasingly closer to becoming a reality. 
Such RL models would follow a similar concept of training a large model on various tasks, environments, and behaviors to be fine-tuned for specific downstream tasks.
SMART~\shortcite{sun-iclr23a}, one of the current contenders for such models, follows this paradigm by using a self-supervised and control-centric objective
that encourages the transformer-based model to capture control-relevant representation and demonstrates superior performance when used for fine-tuning.
AdA~\shortcite{ada-arxiv23a} trains an in-context learning agent on a vast distribution of tasks where the task factors are generated from a latent ruleset. 

Given the pre-training paradigm, these methods are highly data-dependent in principle.
However, incorporating large amounts of data can demonstrate scalability benefits by reducing fine-tuning costs for distributed applications. 
A natural question is the role of Structured RL in the realm of end-end learning and big data.
Even though such methods subscribe to an end-to-end paradigm, structural decompositions can benefit them differently.

\paragraph{Discovering Structure using Foundation Models}
Foundation models offer avenues to discover structure in an unsupervised manner.
Such methods have traditionally been prevalent in unsupervised RL, particularly by learning unsupervised skills. 
Integrating foundation models allows for learning predictive representations of different pipeline parts and discovering behaviors across modalities in different domains. 
Therefore, such methods can offer a good way to implicitly learn decompositions in problems and, consequently, utilize them using the aforementioned patterns.

\paragraph{Interpretability and Selection during Fine-tuning.} 
Researchers can better understand the decomposability in the pre-trained models by categorizing methods based on how they incorporate structure. 
Consequently, this can guide the selection processes for fine-tuning methods depending on the tasks. 
Passive learning from pre-trained models can benefit from better explanations about what parts of a fine-tuning task space might be suited for what kind of warmstarting strategies. Additionally, incorporating interpretability-oriented decompositions such as relational representations can help design more interpretable fine-tuning methods.

\paragraph{Task-Specific Architectures and Algorithms.} 
% Structural information can guide the development of novel architectures.
With a better understanding of how different architectures and algorithms incorporate structural information, practitioners can more effectively adapt existing methods or contribute to designing novel solutions tailored to their specific tasks.
For example, Action Branching architectures might provide modular functionality in downstream tasks, especially suited for multi-task settings. 
On the other hand, representation bottlenecks might suit settings that deviate from each other by small changes in contextual features.

\paragraph{Improved Fine-Tuning and Transfer Learning.} 
By understanding how to decompose tasks and incorporate structural information, foundation models can be fine-tuned more effectively for specific tasks. 
Understanding decompositions could guide how to structure the fine-tuning process or adapt the model to a new task. 

\paragraph{Benchmarking and Evaluation.}
Problems with varying levels of decomposition might have specific benchmarking requirements.
By accounting for these attributes, we can build better benchmarks and evaluation protocols for methods that utilize foundation models.  
Researchers can use this framework to design better evaluation protocols and benchmarks for foundational models. 
For practitioners, such benchmarks and evaluation protocols can guide the selection of models and algorithms for their specific tasks.

\begin{mybox}
\textbf{Research Area 3: Structure and Foundation Models for RL}
\begin{itemize}
    \item Use foundation models to discover decompositions in problems that can be subsequently used for fine-tuning and adaptation.
    \item Use Factored or Relational abstractions on the pre-trained foundation model for state abstractions to manage high-dimensional state spaces and, thus, reduce data dependency.
    \item Condition the policy on additional task-specific information, such as goal information, representation of specific fine-tuning instructions, or control priors to improve scalability. 
    \item Regularize the fine-tuning process to prevent catastrophic forgetting of useful features learned during pre-training.
    \item Maintain a portfolio of fine-tuned policies and value functions to help reuse previously learned skills and adapt them to new tasks, improving learning efficiency and generalization.
    \item Incorporate a curriculum of increasingly complex fine-tuning environments based on the agent's performance to help the agent gradually adapt the foundation model's knowledge to the specific RL task.
    \item Use explicit architectures that fine-tune different RL problem aspects, such as perception, policy learning, and value estimation.
\end{itemize}
\end{mybox}

\subsection{Partial Observability and Big Worlds}
In many real-world situations, the Markov property might not fully capture the dynamics of the environment~\shortcite{whitehead-ai95a,cheung-icml20a}. 
This can happen in cases where the environment's state or the rewards depend on more than just the most recent state and action or if the agent cannot fully observe the state of the environment at each time step.
In such situations, methods must deal with non-Markovian dynamics and partial observability.

\paragraph{Abstractions.}
Abstractions can play a crucial role in such situations, where structural decompositions using abstraction patterns can make methods more sample-efficient. 
Often used in options, temporal Abstractions allow the agent to make decisions over extended periods, thereby encapsulating potential temporal dependencies within these extended actions. 
This can effectively convert a non-Markovian problem into a Markovian one at the level of options.
State abstractions abstract away irrelevant aspects of the state and, thus, can sometimes ignore specific temporal dependencies, rendering the Markovian process at the level of the abstracted states.
Thus, research into the role of decompositions in abstraction opens up possibilities to understand the dependencies between non-Markovian models and the abstractions they use to solve problems with incomplete information.
Abstraction can also simplify the observation space in POMDPs, reducing the complexity of the belief update process. 
The abstraction might involve grouping similar observations, identifying higher-level features, or other simplifications.
Abstractions allow us to break partial observability into different types instead of always assuming the worst-case scenario.
Utilizing such restricted assumptions on partial observability can help us build more specific algorithms and derive convergence and optimality guarantees for such scenarios.
% The agent may need to incorporate additional information into its decision-making processes, such as its belief state or memory of past observations. 
% The augmentation pattern can facilitate this by conditioning the policy on this additional information.

\paragraph{Augmentations.}
Any additional information required, such as belief states or memories of past observations, can be used as abstractions or augmentations.
This can also help with more efficient learning of transition models for planning. 
Hierarchical techniques that utilize optimization at different timescales can incorporate portfolios to reuse learned policies across various levels of abstraction.
Environment generation patterns could also generate a curriculum of increasingly complex tasks for the agent, starting with simpler MDPs and gradually introducing partial observability or other non-Markovian features.

\paragraph{Big worlds.}
As we extend the information content of the environment to its extremity, we delve into the realm of the big world hypothesis in RL~\shortcite{javed-misc23a}, where the agent's environment is multiple orders of magnitude larger than the agent. 
The agent cannot represent the optimal value function and policy even in the limit of infinite data.
In such scenarios, the agent must make decisions under significant uncertainty, which presents several challenges, including exploration, generalization, and efficient learning.
Even though the hypothesis suggests that incorporating side information might not be beneficial in learning the optimal policy and value in such scenarios, structural decomposition of large environments in different ways can allow benchmarking methods along different axes while allowing a deeper study into the performance of algorithms on parts of the environment that the agent has not yet experienced. 
Modular decomposition can guide the agent's exploration process by helping the agent explore different parts of the environment independently. 
Incorporating modularity opens a gateway to novel methods and theoretical insights about the relationships between task decomposition, exploration, and learning efficiency in large environments.
Relational decompositions can help the agent learn relationships between different entities, bolstering its ability to generalize to unseen parts of the environment. 
Finally, Structural information can be used to facilitate more efficient learning. For instance, in an auxiliary optimization pattern, the agent could learn faster by optimizing auxiliary tasks that are easier to learn or provide helpful information about the environment's structure.

\begin{mybox}
    \textbf{Research Area 4: Structure in Partial Observability and Big Worlds}
    \begin{itemize}
        \item Use temporal and state abstraction to abstract away temporal dependencies and non-Markovian aspects of the state. Utilize Modularity to tie these abstractions to learned primitives such as skills or options.
        \item Use memory more efficiently as an abstraction or augmentation for learned transition models.
        \item Create a portfolio of policies and utilize them for optimization across timescales, such as in hierarchical methods, to make them more tractable.
        \item Utilize modular decompositions for guiding separate and parallel exploration mechanisms for different parts of the state space. Utilize relational abstractions to make this knowledge more interpretable.
        \item Utilize structure for task factorization to guide benchmarking methods along different axes of task complexity.    
    \end{itemize}
\end{mybox}

\subsection{Automated Reinforcement Learning (AutoRL)}
Automated RL (AutoRL) is a sub-field focused on methods to automate the process of designing and optimizing RL algorithms, including the agent's architecture, reward function, and other hyperparameters~\shortcite{parkerholder-jair22a}. 
Methods in AutoRL can be placed on a spectrum of automation, where on one end would be methods to select pipelines and on the other would be methods that try to discover new algorithms ground-up in a data-driven manner~\shortcite{oh-neurips20a}.
Techniques from the Automated Machine Learning literature~\shortcite{hutter-book19a} then transfer to the RL setting, including algorithm selection~\shortcite{laroche-iclr18a}, hyperparameter optimization~\shortcite{li-kdd19a,parkerholder-neurips20a,wan-automl22a}, dynamic configurations~\shortcite{adriaensen-jair22a}, learned optimizers~\shortcite{lan-arxiv23a}, and neural architecture search~\shortcite{wan-automl22a}. 
Similarly, techniques from the Evolutionary optimization and Meta-Learning literature naturally transfer to this setting with methods aiming to meta-learn parts of the RL pipeline such as update rules~\shortcite{oh-neurips20a}, loss functions~\shortcite{salimans-arxiv17b,kirsch-iclr20a}, symbolic representations of algorithms~\shortcite{alet-iclr20a,coreyes-iclr21b,luis-arxiv22a}, or concept drift~\shortcite{lu-neurips22a}.
However, there are still many open questions in AutoRL, such as properties of hyperparameter landscapes in RL~\shortcite{mohan-automlconf23a}, sound evaluation protocols~\shortcite{eimer-icml23a}, stability of training due to the non-stationary learning task and non-deterministic data collection on the fly. 
Consequently, most of these methods suffer from a lack of scalability. 

\paragraph{Algorithm Selection and Configuration.} 
Depending on the decomposability of the problem at hand, different RL methods could be more appropriate. 
Structural decompositions can guide the selection process in AutoRL by suggesting appropriate types of decompositions based on the problem characteristics. 
Understanding how different decomposition types influence the performance of RL methods can bridge the gap between selection and configuration by helping researchers understand the level of abstraction needed for selection conditioned on the task, aiding in developing more efficient and targeted search algorithms.
Decomposability can also guide ranking procedures, where methods that cater to different decomposability can be ranked differently,  given a problem.

\paragraph{Hyperparameter Optimization.} 
Parameters related to structural decomposition (e.g., the number of subtasks in a modular decomposition) could be part of the hyperparameter optimization process in AutoRL. 
Researchers can investigate the interplay between configuration spaces of hyperparameters and various structural decomposition-related parameters. 
For example, a high decomposability problem might require different exploration or learning rates than a low one.
This could lead to novel insights and methods for more effective hyperparameter optimization in AutoRL. 
Practitioners can use this understanding to guide the hyperparameter optimization process in their AutoRL system. 
By tuning parameters related to the decomposition, they can potentially improve the performance of their RL agent.

\begin{mybox}
    \textbf{Research Area 5: Structure in AutoRL}
    \begin{itemize}
        \item Use methods to perform a decomposability assessment of a problem. This can help guide algorithm selection for problems with different types of decomposability.
        \item Expedite the hyperparameter search by abstracting away task-irrelevant aspects.
        \item reuse learned policy and value functions stored in a portfolio for landmarking performances of algorithms similar to the one being optimized.
        \item Incorporate modularity information in the form of goals and task hierarchies in the search process. 
        \item Structure neural architecture search-space using decomposability in the problem.
    \end{itemize}
\end{mybox}

\subsection{Meta-Reinforcement Learning}

Meta-reinforcement learning, while having overlaps with AutoRL, is a field in and of itself~\shortcite{beck-arxiv23a} that focuses on training agents to adapt and learn new tasks or environments quickly.
The general Meta-RL setup involves a bi-level optimization procedure where an agent learns a set of parameters by training on a distribution of tasks or environments that help it adapt and perform well on new, unseen tasks that share some form of overlap with the training tasks.
\shortciteA{beck-arxiv23a} outline different problem settings in Meta-RL based on the kind of feedback (supervised, unsupervised, rewards) provided to the agent during the training and adaptation phases.
We mainly refer to the standard setting where extrinsic rewards act as feedback during the training and adaptation phases. 
However, decompositions can also be helpful for settings like those discussed in other sections.

\paragraph{Task Decompositions.} 
Different task decomposition approaches could guide the meta-learning process depending on the meta-task's decomposability. 
Consequently, understanding how task decomposition affects Meta-RL can guide the development of more effective meta-learning algorithms. 
It might also lead to new insights on balancing learning between different subtasks. 
By identifying suitable decompositions, practitioners can set up their system to learn in a way that is more aligned with the structure of the tasks, potentially leading to improved performance.

\paragraph{Adaptation Strategies.} 
Decompositions could inform the way a Meta-RL agent adapts to a new task. 
For instance, if the new task is highly decomposable, a modular adaptation strategy could be more appropriate by guiding the agent to an appropriate latent space of the new task. 
Thus, incorporating existing patterns can inspire new research into how the task's decomposition can guide adaptation strategies in Meta-RL. 
This could lead to novel methods or theories on adapting to new tasks more effectively based on their structure.

\begin{mybox}
    \textbf{Research Area 6: Structured Meta-RL}
    \begin{itemize}
        \item Use decompositions for abstracting task distributions, which can be integrated into the adaptation process.
        \item Compartmentalize the learning process into modules for highly decomposable problems. These modules can serve as abstract configurations for the meta-level and, thus, make the outer loop more tractable 
        \item Learn and create a portfolio of models geared towards specific task clusters for different decomposability types to guide data augmentation during the adaptation phase.
        \item Utilize decomposability to design learning curricula based on abstract types of tasks to train the warmstarting configurations
    \end{itemize}
\end{mybox}

% \section{Discussion and Future work}
% \label{sec:takeaways}

% Designing RL pipelines that can generalize to new tasks while being able to perform well on specific problems requires carefully understanding the balance between these two ends of the dichotomy. 
% Structural decompositions and patterns of incorporating them provide a novel way to understand how to address problems, with biases of each pattern already illuminating the techniques suited for different objectives.
% Patterns such as abstraction, augmentation, and warehousing lend themselves well to generalization. 
% Abstraction and augmentation, for example, can create more generalized state or action representations or integrate additional information into the learning process, enabling the agent to handle varying dynamics. 
% Warehousing, meanwhile, promotes the storage and reuse of learned policies or value functions, enhancing the agent's ability to adjust to new tasks.
% Patterns such as auxiliary optimization, environment generation, and explicitly designed architecture mainly target more deployability-suited objectives.
% Auxiliary optimization can incorporate task-specific constraints or objectives, while environment generation can create a task sequence that progressively leads to the target task. 
% Explicitly designed architectures, on the other hand, can help tailor the learning process to specific tasks or environments.

% However, this is not a conclusive statement, as we see ample works that combine patterns to address either objective.
% Additionally

\section{Conclusion and Future Work}

% We introduce a novel framework of different patterns for incorporating the structure of a learning problem as an inductive bias into RL algorithms. 
% We first ground structure as side information about decomposability in a learning problem and potential solutions.
% By categorizing decomposability into four archetypes along a spectrum, we establish connections with existing literature, shedding light on how structure influences RL.
% Through a meticulous analysis of the RL landscape, we identify seven patterns that serve as robust pathways for integrating structural knowledge. 
% % These patterns provide a repeatable framework for incorporating side information biases into the learning pipeline, enhancing RL performance and capabilities.
% Our research concludes with a pattern-centric lens, revealing the vital role of structural decompositions in present and future RL paradigms. 
% We aim to inspire researchers and practitioners to embrace this perspective, fostering advancements and innovation in the field of RL.
% By presenting this comprehensive framework, we provide a valuable resource for researchers, facilitating further exploration and investigation into the incorporation of structure in RL.

Understanding the intricacies and complexities of Deep RL is challenging, exacerbated by the divergent methodologies employed across different problem domains. 
This fragmentation hinders the development of unifying principles and consistent practices in RL. 
To address this critical gap, we propose an innovative framework to understand different methods of effectively integrating the inherent structure of learning problems into RL algorithms. 
Our work serves as a pivotal step towards consolidating the multifaceted aspects of RL, ushering in a design pattern perspective for this domain.

We first conceptualized structure as side information about the decomposability of a learning problem and corresponding solutions. 
We have categorized decomposability into four distinct archetypes - latent, factored, relational, and modular. 
This classification delineates a spectrum that establishes insightful connections with existing literature, elucidating the diverse influence of structure within RL.

We then presented seven key patterns following a thorough analysis of the RL landscape - abstraction, augmentation, auxiliary optimization, auxiliary model, warehousing, environment generation, and explicitly designed patterns. 
These patterns represent strategic approaches for the incorporation of structural knowledge into RL. 
Although our framework provides a comprehensive starting point, we acknowledge that these patterns are not exhaustive. 
We envisage this as an impetus for researchers to refine and develop new patterns, thereby expanding the repertoire of design patterns in RL.

In conclusion, our work offers a pattern-centric perspective on RL, underlining the critical role of structural decompositions in shaping both present and future paradigms. 
By promoting this perspective, we aim to stimulate a new wave of research in RL, enriched by a deeper and more structured understanding of the field. 
While our proposed framework is a novel contribution, it should be viewed as an initial step in an ongoing process. 
We anticipate and encourage further development and refinement of our framework and eagerly await the emergence of new, innovative patterns that will undoubtedly shape the future of RL.

% \section{Acknowledgements}
% We want to thank Robert Kirk and Rohan Chitnis for their discussion and comments on drafts of this work.
\newpage
\acks{We wish to thank Robert Kirk and Rohan Chitnis for their discussion and comments on drafts of this work. We would also like to thank Vincent François-Lavet, Khimya Khetarpal, and Rishabh Aggarwal for providing additional relevant references in the literature.
}

% \newpage

\vskip 0.2in
\typeout{}
\bibliography{bib/strings, bib/lib, bib/local, bib/patterns, bib/types, bib/proc}
\bibliographystyle{theapa}
% \printbibliography 

\end{document}